\documentclass{article} %
\usepackage[final]{colm2026_conference}

\usepackage{microtype}
\usepackage{hyperref}
\usepackage{url}
\usepackage{booktabs}

\usepackage[utf8]{inputenc}
\usepackage[T1]{fontenc}
\usepackage{amsmath,amssymb,amsfonts}
\usepackage{pifont}
\usepackage{graphicx}
\usepackage{booktabs}
\usepackage{hyperref}
\usepackage{xcolor}
\usepackage{enumitem}
\usepackage{caption}
\usepackage{subcaption}
\usepackage{natbib}
\usepackage{microtype}
\usepackage{tabularx}
\usepackage{multirow}
\usepackage{float}

\usepackage{xcolor}
\usepackage{soul}

\usepackage{lineno}

\definecolor{darkblue}{rgb}{0, 0, 0.5}
\hypersetup{colorlinks=true, citecolor=darkblue, linkcolor=darkblue, urlcolor=darkblue}

\title{Spoiler Alert: Narrative Forecasting as a Metric for Tension \\ in LLM Storytelling}

\author{
\parbox{\linewidth}{
Peiqi Sui\textsuperscript{1}\thanks{Equal contribution.}, ~~Yutong  Zhu\textsuperscript{2}\footnotemark[1],\\ 
~~Tianyi Cheng\textsuperscript{2}, ~~Peter West\textsuperscript{3}, ~~Richard Jean So\textsuperscript{4}, ~~Hoyt Long\textsuperscript{2}, ~~Ari Holtzman\textsuperscript{2}
}\\
\textsuperscript{1}McGill University\hspace{5.7pt}
\textsuperscript{2}University of Chicago\hspace{5.7pt}\\
\textsuperscript{3}University of British Columbia\hspace{5.7pt}
\textsuperscript{4}Duke University \\
\texttt{peiqi.sui@mail.mcgill.ca}, \texttt{\{maxzhuyt,hoytlong,aholtzman\}@uchicago.edu} \\
}

\begin{document}

\ifcolmsubmission
\linenumbers
\fi

\maketitle

\begin{abstract}

LLMs have so far failed both to generate consistently compelling stories and to recognize this failure—on the leading creative-writing benchmark (EQ-Bench), LLM judges rank zero-shot AI stories above \textit{New Yorker} short stories, a gold standard for literary fiction. We argue that existing rubrics overlook a key dimension of compelling human stories: narrative tension. We introduce the \textit{100-Endings} metric, which walks through a story sentence by sentence: at each position, a model predicts how the story will end 100 times given only the text so far, and we measure tension as how often predictions fail to match the ground truth. Beyond the mismatch rate, the sentence-level curve also yields complementary statistics that track plot-level twists and revelations, such as the \textit{inflection rate}, a geometric measure of how frequently the curve reverses direction. Unlike rubric-based judges, \textit{100-Endings} correctly ranks \textit{New Yorker} stories far above LLM outputs. Grounded in narratological principles, we design a story-generation pipeline using structural constraints, including analysis of story templates, idea formulation, and narrative scaffolding. Our pipeline significantly increases narrative tension as measured by the \textit{100-Endings} metric, while maintaining performance on the EQ-Bench leaderboard.

\end{abstract}

\begin{figure*}[!ht]
    \centering
    \begin{minipage}{0.70\textwidth}
        \centering
        \includegraphics[width=\textwidth]{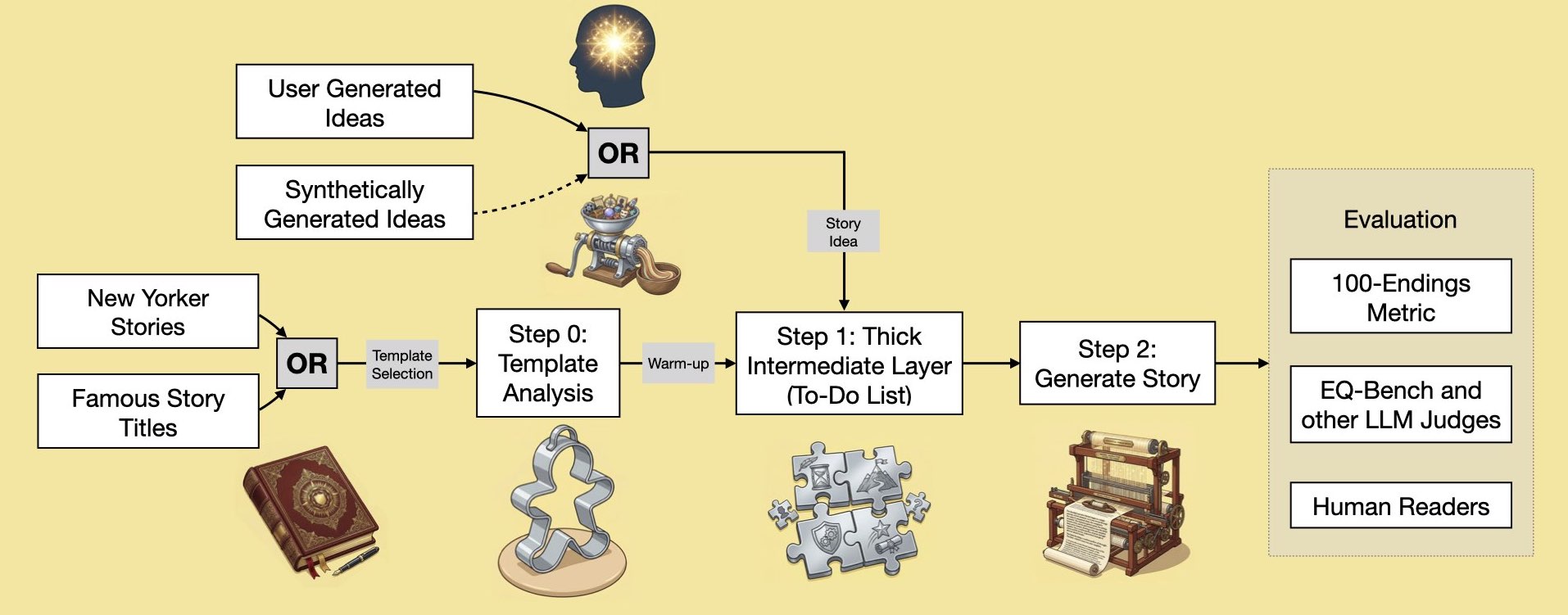}
    \end{minipage}
    \hfill
    \begin{minipage}{0.29\textwidth}
        \centering
        \includegraphics[width=\textwidth]{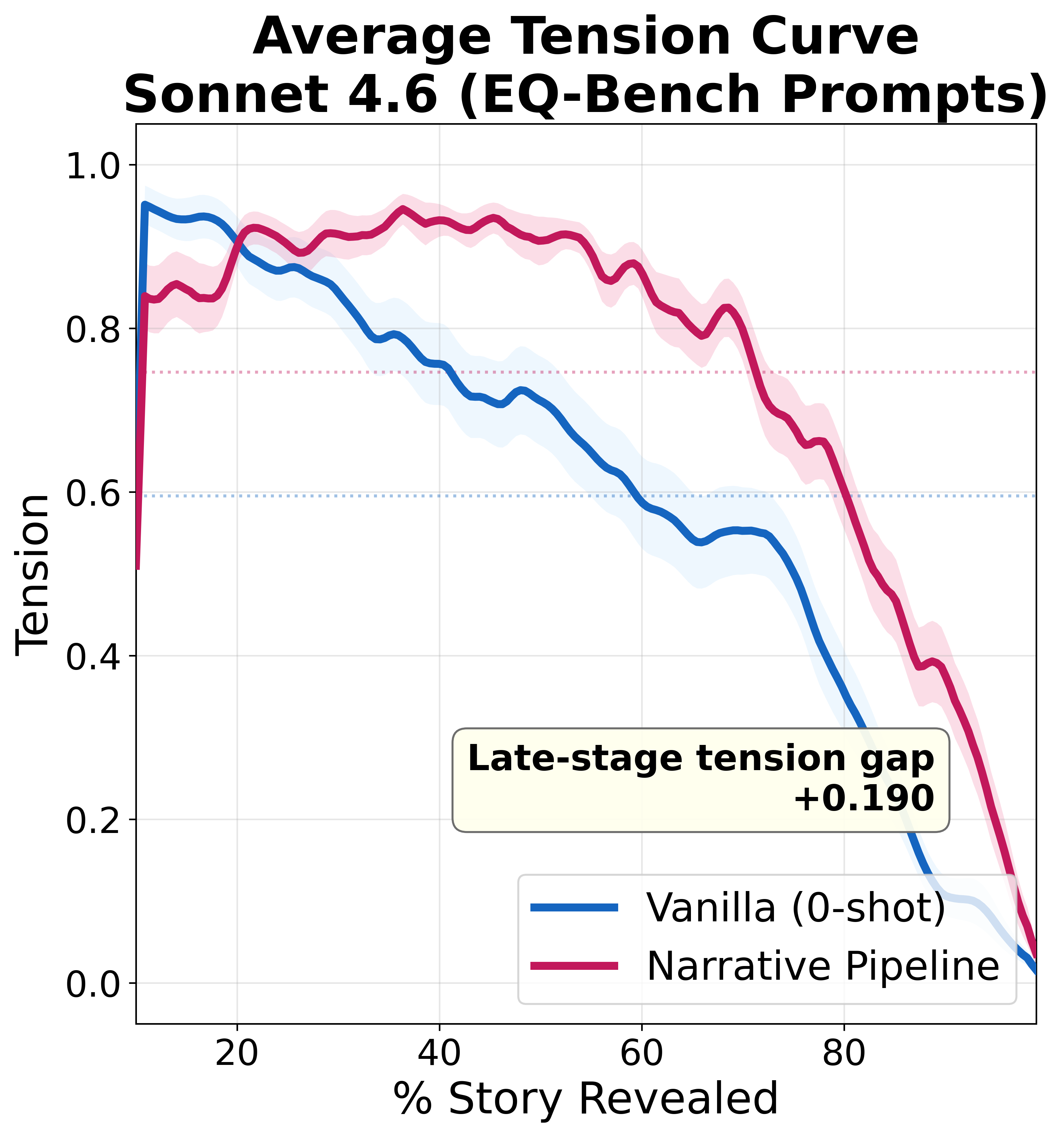}
    \end{minipage}

    \caption{Pipeline architecture and effects. \textbf{Left:} the pipeline extracts a beat template from a reference story (Step~0), combines it with a target idea into a thick intermediate plan, i.e.\ a densely structured beat scaffold (Step~1), and generates the story (Step~2). \textbf{Right:} effect on the shape of \textit{100-Endings} tension curves.}
    \label{fig:flow_chart}
\end{figure*}

\section{Introduction}

Even the largest and strongest LLMs have trouble with creativity: 
AI writers struggle to avoid clichés \citep{chakrabarty2024art}, produce diverse plot and narrative arcs \citep{padmakumar2024diversity,xu2025echoes}, or maintain long-range narrative coherence \citep{gurung2025learning,pham2025frankentext}. This failure extends to evaluation, with strong, widely-adopted model judges rating professional human-written stories below low-quality AI generations (Section~\ref{sec3.2}). Increasing model scale does not resolve these shortcomings, which indicates that core aspects of compelling stories may not be naturally understood by zero-shot LLMs. We investigate how these attributes can be explicitly incorporated into current LLM evaluators and writers, focusing on \textbf{narrative tension} as a key signpost for quality in human-written stories.

On their own, LLMs regularly fail to recognize their lack of narrative tension in writing. Humans often note that LLM stories are flat and trite \citep{chakrabarty2025salvaged}, while model judges typically focus on other aspects of their writing. Most rubrics simply omit tension as an evaluation criterion, but even when it is included, model evaluators tend to reward surface markers of drama like urgent language or explicit emotional labels, rather than verifying that information is genuinely withheld across the narrative arc. More fundamentally, rubric-based evaluation assumes that story quality is an average over item scores, yet this is not how human readers experience fiction: readers do not consult a checklist; they read to be immersed, to feel uncertain and compelled, to arrive at the ending slightly transformed. This visceral sense of tension is immediate and holistic, and a key component of what makes a story worth reading—yet no rubric item, however well-worded, gives a model judge access to it. Reliable measurement therefore requires an approach that operationalizes tension structurally rather than delegating it to a judge's proxy.

To address this gap, we introduce the \textit{100-Endings} metric that directly targets narrative tension \emph{by design}. Following insights from literary theory (Section~\ref{sec2.1}), we operationalize tension as unpredictability: after every sentence of a story, a model is asked to infer 100 possible endings, and we use their divergence from the ground truth ending as a measurement for tension. \textit{100-Endings} flags predictable stories when the model successfully anticipates their endings, while stories with genuine tension would exhibit narrative recalcitrance that resists easy predictions. Unlike the baseline of standard evaluation practices, our \textit{100-Endings} correctly identifies the high tension of high-quality stories from \textit{The New Yorker}, widely regarded as a gold standard of contemporary literary fiction (Section~\ref{sec3}).

Using \textit{100-Endings} as a north star, we investigate what needs to be injected into LLMs to write high-tension stories. At a high level, the answer seems to be the complexity of inputs and structural constraints. We design a generation pipeline grounded in literary theory that can take a flexible range of story ideas and transform them into tension-rich stories through a thick intermediate plan: a dense beat-level scaffold specifying narrative function, information withholding, and tension mechanics, derived from an analysis of professional fiction (Section~\ref{sec4}). This pipeline closes roughly half the gap to professional \textit{New Yorker} fiction on our metric while achieving the top score on the EQ-Bench Creative Writing leaderboard (Section~\ref{sec5}). Our contributions are as follows:

\begin{enumerate}
    \item \textbf{A new metric for narrative tension.} We introduce the \textit{100-Endings Metric}, which operationalizes tension as unpredictability of story endings at the sentence level—capturing a crucial dimension of story quality missed by existing rubrics (Section~\ref{sec3.1}).
    \item \textbf{Evidence that LLMs systematically cannot hold tension.} Using 100-Endings, we show that top LLMs produce stories with the lowest narrative tension of any corpus we evaluate, even as rubric-based judges rate them above professional human fiction (Section~\ref{sec3.2}).
    \item \textbf{A structured pipeline improves tension.} We propose a story generation pipeline grounded in literary theory and show that structured constraints improve narrative tension (Section~\ref{sec5}).
\end{enumerate}

This work highlights the need for deliberate, literary theory-driven solutions for LLM creativity: scale alone has failed to capture fundamental narrative elements such as tension, while explicitly modeling it creates significant and demonstrable improvements. Tension is by no means the only missing element in LLM writers, and explicit, narratology-driven work will be needed to map out and implement the full space of key narrative elements needed for true narrative machines.

\section{Related Work}
\label{sec2}

\subsection{Literary and Narrative Theories of Tension.}
\label{sec2.1}
Narratology has long considered tension, an affective state driven by information asymmetry and delayed resolution, as central to storytelling. Tension comes from the structural divide between a story's chronological events (\textit{fabula}) and the reconfigured order in which they are disclosed (\textit{syuzhet}) \citep{propp1928morphology}, and sustaining that divide requires restricted focalization, perspectives that deliberately limit the information available to the reader at any moment \citep{genette1980narrative}. Tension operates through the active management of narrative expectations, notably suspense generated by forward-looking uncertainty about unresolved outcomes \citep{sternberg2003universals}. Structurally, it depends on open, high-stakes questions posed early and left unanswered until the ending, forming causal chains whose strength rests on the viability of competing answers \citep{carroll1990philosophy}; conversely, if the resolution is disclosed prematurely, this dynamic collapses and tension vanishes. As an affect, tension integrates uncertainty with anticipation, which requires compelling plots to continuously deviate from the expected norm  to generate ``eventfulness'' \citep{baroni2007tension}. 

These formulations find empirical support in cognitive science: narrative structure reliably produces distinct affective states by temporally separating initiating events from outcomes, forcing readers into sustained predictive inference \citep{brewer1982stories,tobin_elements_2018}; the greater the divergence between possible fictional outcomes, the higher the reader's experienced tension \citep{lehne2015reading}. Beyond the literary domain, this friction demonstrates broader social affordances as a reaction to cognitive dissonance \citep{lehne2015tension}, where tension is more generally formalized as the variance in beliefs about future outcomes \citep{ely2015suspense}. 

Computational approaches to suspense have not produced a generalizable, automated metric for evaluating tension as a story unfolds. \citet{oneill2014dramatis} count a protagonist's available escape plans and success probabilities, but require a symbolic planner with hand-engineered action spaces. \citet{wilmot2020modelling} model suspense as uncertainty reduction over neural story representations, correlating strongly with human ratings but requiring human annotations. More recently, \citet{tian2024narratives} tracks tension via word-level arousal curves, \citet{xie2024suspenseful} use human pairwise judgments, and \citet{matlin2025languagemodelsagreehuman}  show LLMs cannot reliably quantify suspense. Building on these accounts, we operationalize narrative tension through \emph{predictability of endings} (Section~\ref{sec3.1}), a structural prerequisite for tension rather than a proxy of self-reported ratings.

\subsection{Evaluation of LLM Creative Writing.}
\label{sec2.2}

Scalable evaluation of LLM fiction has converged on the LLM-as-a-judge paradigm \citep{zheng2023judging,liu2023geval,kim2024prometheus} that scores outputs against multi-criteria rubrics. The widely adopted EQ-Bench \citep{Paech2023}, for example, evaluates 22 criteria like prose quality and character authenticity. While these rubrics can separate polished prose from poor writing, they struggle to capture deep structural mechanics that are representative of more nuanced aspects of literary quality \citep{chakrabarty2024art,feuer2025style}. Crucially, direct measures of narrative tension are absent from them altogether. This gap is further exacerbated by the self-preference bias of LLM judges, which favors the resolution-oriented, affectively flat styles that aligned models often produce \citep{panickssery2024llm}.

Beyond rubrics, automated metrics for narrative quality have progressed from reference-dependent overlap measures, which are largely inapplicable to open-ended generation~\citep{guan2020union,chhun2022human}, toward structural approaches like coherence-oriented frameworks \citep{wang2025characterbox,yi2025score} and information-theoretic methods that quantify the predictability gap between human and LLM writing \citep{xu2025echoes,sui2026llmsexhibit}. However, these metrics evaluate predictability as an artifact of raw continuation, capturing whether the next sentence is surprising but not whether the story's ultimate resolution remains uncertain throughout; a story may exhibit high local novelty while revealing its ending early, producing superficial surprise without sustained tension. \textit{100-Endings} targets this gap, shifting the unit of evaluation from immediate next-step continuation to the story's resolution and measuring forecasting difficulty across the full narrative arc.

\section{The \textit{100-Endings} Metric}
\label{sec3}

\begin{figure*}
    \centering
    \begin{minipage}[t]{0.42\textwidth}
        \centering
        \vspace{0pt}
        \includegraphics[width=\linewidth]{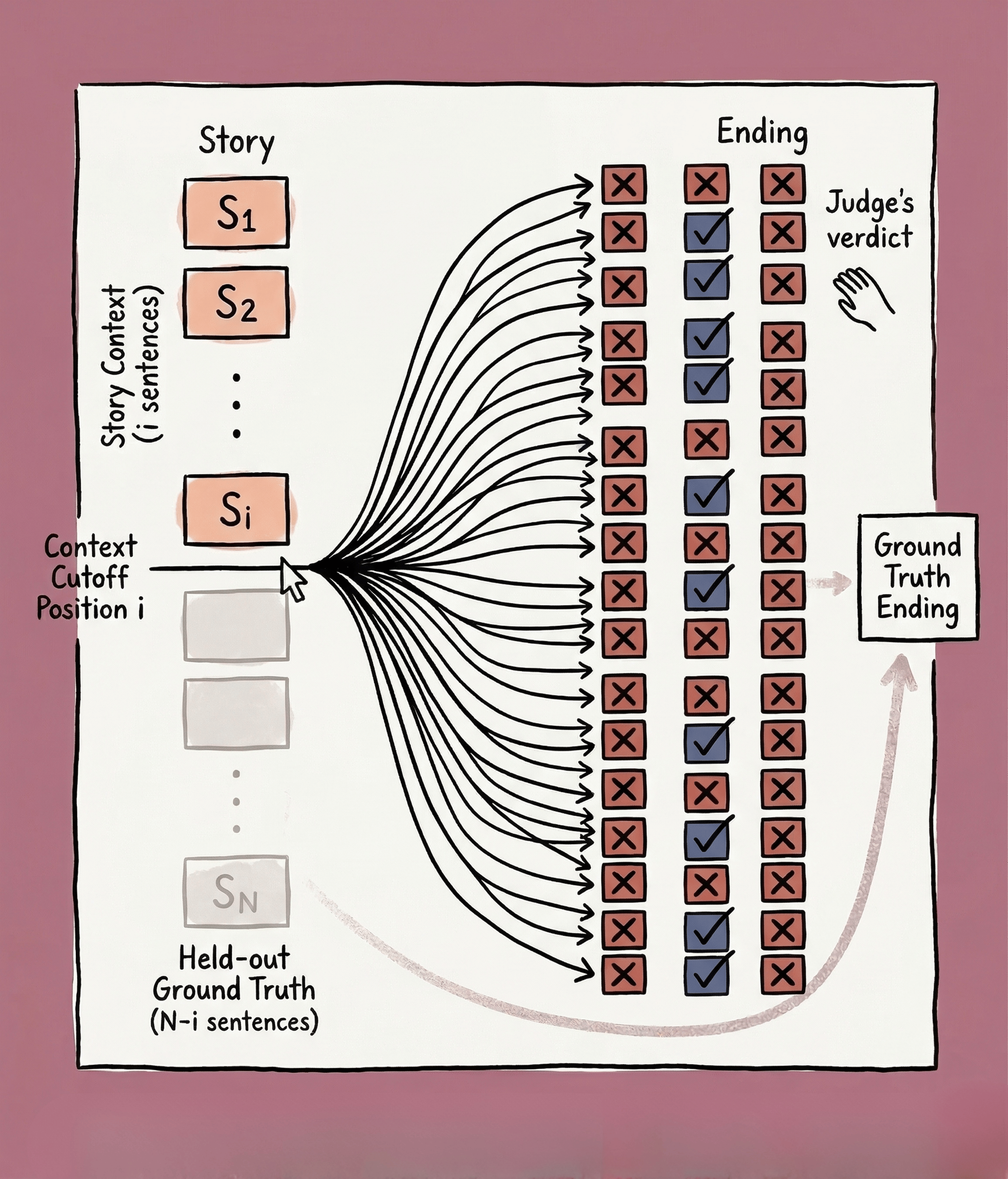}
    \end{minipage}%
    \hfill
    \begin{minipage}[t]{0.54\textwidth}
        \vspace{0pt}
        \small
        \textbf{Example: ``Snowing in Greenwich Village''}\\[3pt]
        \textit{Richard walks Rebecca home through the snow. She invites him up to her tiny apartment under a slanting roof. He sees the bed, the stove on the fridge; they look out the window...}\\[6pt]
        \textbf{Sentence 198 (91.2\% revealed):}\\
        \textit{``Well''---this word too loud; he finished the sentence too softly---``thanks for letting me see it.''}\\[4pt]
        \colorbox{green!8}{\parbox{0.96\linewidth}{%
        \textbf{\textcolor{green!50!black}{\ding{51}~Matched:}}
        ``The quiet between them thickened, neither stepping forward nor back, and just as the snow seemed to settle over the city like a layer of pure silence, Richard reached out---not to embrace or undo, but merely to touch the curtain, letting the folds fall gently back into place between them.''}}\\[4pt]
        \colorbox{red!6}{\parbox{0.96\linewidth}{%
        \textbf{\textcolor{red!60!black}{\ding{55}~Diverged:}}
        ``They stood a moment longer in the windowed silence of her small room, snow falling beyond the pane, and then Richard reached out, gently, and pulled her to him. She did not resist, but neither did she respond.''}}
    \end{minipage}
    \caption{\textbf{Left}: Illustration of the \textit{100-Endings} metric. At each sentence position, 100 candidate endings are generated and judged against the actual continuation; the fraction that diverges (\textit{no-rate}) tracks narrative tension. \textbf{Right}: Example from Updike's ``Snowing in Greenwich Village'' at 91.2\% revealed (no-rate = 0.45). 
    }
    \label{fig:metric}
\end{figure*}

We introduce the \textit{100-Endings} metric (Figure~\ref{fig:metric}), which measures how predictable a story's ending is at each point in its progression. At every sentence boundary, we prompt a language model to predict how the story will end given only the text so far, then compare those predictions against the held-out ground truth ending. The core intuition is that tension is operationally linked to predictability: a story that sustains information asymmetry, keeps possibilities open, and delays resolution should resist easy prediction and allow divergent trajectories that depart from the actual ending; a story that lacks tension will be easy to anticipate even from partial text. Unlike adjacent approaches that measure the predictability of the immediate next sentence \citep{xu2025echoes,sui2026llmsexhibit}, our metric evaluates the forward-looking narrative trajectory with a focus on how the story ends. Despite the name, \textit{100-Endings} is not a metric \emph{of} the ending: the ending serves as a fixed target against which we measure how much the story has already given away, and the resulting statistics describe tension across the entire arc.

\subsection{Evaluation Procedure}
\label{sec3.1}

\paragraph{Ending Generation.}
Given a story of $N$ sentences, we evaluate sentence ending positions $P$ spanning 10\%--99\% of the text of the story.\footnote{Positions below 10\% provide too little context for meaningful prediction; sentences under 10 words are concatenated to the next.} At each position $i \in P$, we provide sentences $1 \ldots i$ as context to prompt a generation model (\textsc{Qwen3-32B}, $T=1.2$) to bypass intermediate plot developments and explicitly produce 100 candidate endings. The 100 endings at each position are independent generations, sampled separately instead of parsed from a single model call; Appendix~\ref{app:ending-diversity} confirms the endings are genuinely varied and the no-rate is not affected by lexical diversity.

\paragraph{LLM-as-a-Judge Ending Evaluations.}
A separate judge model (\textsc{Qwen3-8B}, $T=0$) then evaluates each of the 100 generated endings against the entirety of the held-out ground truth (sentences $i{+}1 \ldots N$ of the original story). For each ending, the judge determines whether the plot, events, or actions of the generated ending overlap with the actual ending, following a strict standard where vague thematic similarity is not sufficient. 

The judge returns a binary verdict (Y/N) for all 100 endings in a single pass. We define the \textbf{no-rate} at position $i$ as the fraction of 100
endings that do \emph{not} match the actual continuation:
\begin{equation}
\label{eq:norate}
\text{no-rate}(i) = \frac{1}{100} \sum_{j=1}^{100}
    \mathbf{1}\!\left[\,\text{ending}_j \text{ does not match true
    continuation}\,\right]
\end{equation}
A no-rate of $0.85$ means that 85 of 100 generated endings fail to predict what actually happens. In other words, the story up to this point is highly unpredictable.

\paragraph{Mean no-rate.}
We compute the story-level no-rate by averaging across all evaluated
positions, then report the average across all stories in a group:
\begin{equation}
\label{eq:norate-agg}
\overline{\text{no-rate}} = \frac{1}{|S|} \sum_{s \in S}
    \left( \frac{1}{|P_s|} \sum_{i \in P_s} \text{no-rate}_s(i) \right)
\end{equation}
where $S$ is the set of stories and $P_s$ is the set of evaluated positions in story $s$. Mean no-rate is our primary metric throughout this paper.

\paragraph{Inflection rate.}
Because no-rate is computed at every sentence boundary, it also produces a \emph{curve} over the arc of the story. A complementary statistic is the \textbf{inflection rate} IR$_\alpha$: the fraction of evaluated positions at which the smoothed no-rate curve sharply reverses direction, i.e.\ where the vertex angle $\theta_i \leq \alpha$. Angles are computed after rescaling each story's curve onto a unit square, so that thresholds are comparable across stories of different lengths and no-rate ranges; Appendix~\ref{app:inflection-geometry} gives the full derivation. We report IR$_{60}$ throughout.

\begin{figure}[t]
    \centering
    \includegraphics[width=0.62\columnwidth]{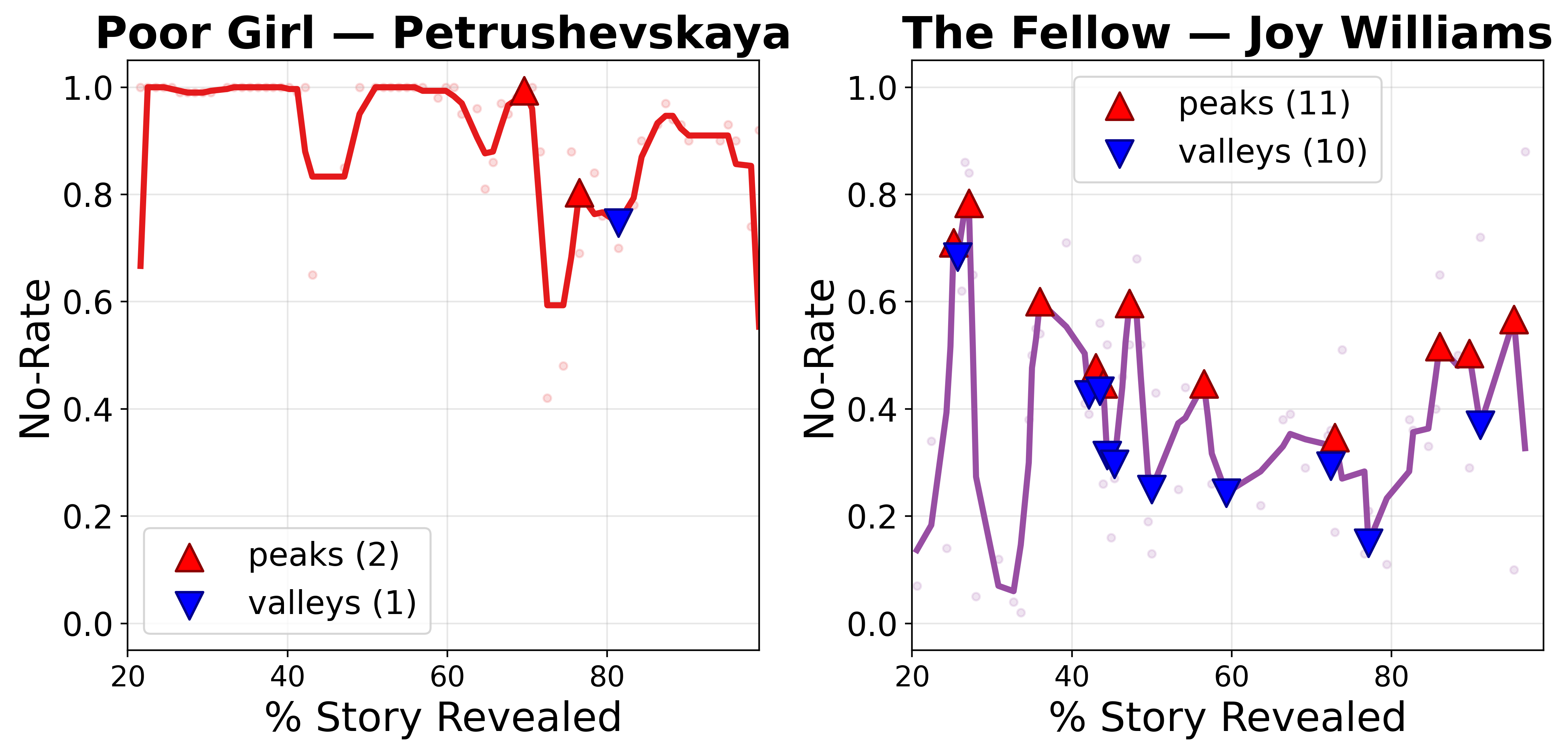}
    \caption{Mean no-rate and inflection rate can diverge. \textbf{Left:} ``Poor Girl'' (Petrushevskaya), very high mean no-rate (0.925) but a nearly flat curve (IR$_{60}$ 0.048). \textbf{Right:} ``The Fellow'' (Joy Williams), low mean no-rate (0.403) but frequent sharp reversals (IR$_{60}$ 0.339). Triangles mark peaks/valleys with vertex angle $\leq 60°$.}
    \label{fig:inflection-examples}
\end{figure}

\noindent While mean no-rate summarizes overall unpredictability, inflection rate tracks narrative manipulation, the frequency with which a story reverses reader expectations. Figure~\ref{fig:inflection-examples} illustrates this with two \emph{New Yorker} stories. Petrushevskaya's ``Poor Girl'', a realist story with a clinical and distanced exteriority, has high mean no-rate ($0.925$) but low inflection rate ($0.048$). ``The Fellow'' (Joy Williams), a surrealist story in which a guest at a writers' residency vanishes in a flood and a dog begins to speak, has low mean no-rate ($0.403$) because the plot is minimal, but very high inflection rate ($0.339$): each small revelation briefly makes the ending guessable before the next twist displaces it. Whether these two statistics capture genuinely distinct dimensions of craft is an open question, but the contrast suggests that the no-rate curve contains information beyond its mean.

\paragraph{Post-spike retention.}
For each local peak in the no-rate curve (a position exceeding both neighbors), we measure how much of the peak value is retained at the minimum within the next 10 positions. The story-level \textbf{retention} is the mean across all peaks. Higher retention means tension peaks are sustained rather than immediately resolved.

\subsection{Why a New Metric?}
\label{sec3.2}
EQ-Bench Creative Writing v3~\citep{Paech2023} is one of the most widely used benchmark for evaluating LLM-generated fiction. Its rubric scores stories on 22 criteria (e.g., adherence to instructions, prose quality, emotional complexity, etc.) using Claude Sonnet 4 as judge, producing a composite score on a 0--100 scale. Notably, none of the 22 directly evaluate for tension.\footnote{While EQ-Bench also employs an Elo rating system based on pairwise matchups, we evaluate only the rubric scores. The leaderboard recently transitioned to Sonnet 4 as a judge (previously Sonnet-3.7). Examination of the repository indicates that while rubric scores have been updated, most Elo scores have not, making pairwise comparisons inconsistent. Furthermore, because the Elo ranking criteria closely mirror the initial rubric and entirely omit narrative tension, the same evaluative limitations apply to both.}

We validate both EQ-Bench and our \textit{100-Endings} metric on four groups of stories of varying quality (details see Appendix): \textit{New Yorker} fiction (professional), Tell Me a Story (amateur, curated), StoryStar (amateur, uncurated), and LLM-generated stories by top performers from the EQ-Bench leaderboard (zero-shot outputs on 32 standard prompts).

\begin{table*}[t]
\centering
\small

\setlength{\tabcolsep}{4pt}

\begin{minipage}[t]{0.4\textwidth}
\centering
\begin{tabular}[t]{lc}
\toprule
\textbf{Source} & \textbf{EQ-Bench} \\
\midrule
GPT-5.2                          & \textbf{83.80} \\
Claude Sonnet 4.6                & 82.04 \\
Claude Opus 4.6                  & 82.88 \\
\midrule
\emph{New Yorker} (professional) & \textbf{78.61} \\
Tell Me a Story (amateur)        & 60.89 \\
StoryStar (amateur)              & 45.81 \\
\bottomrule
\end{tabular}
\end{minipage}\hfill %
\begin{minipage}[t]{0.58\textwidth}
\centering
\begin{tabular}[t]{lcccc}
\toprule
\textbf{Group} & \textbf{Mean} & \textbf{Late} & \textbf{Retain} & \textbf{Infl$_{60}$} \\
\midrule
\emph{New Yorker} & \textbf{0.765} & \textbf{0.607} & \textbf{52\%} & \textbf{0.200} \\
Tell Me a Story   & 0.693 & 0.311 & 39\% & 0.128 \\
StoryStar         & 0.656 & 0.375 & 35\% & 0.166 \\
Top-10 LLMs       & 0.630 & 0.215 & 23\% & 0.106 \\
\bottomrule
\end{tabular}
\end{minipage}
\setlength{\tabcolsep}{6pt}
\vspace{2mm}
\caption{EQ-Bench rubric score (left) vs.\ \textit{100-Endings} (right). \textbf{Mean}: story-level no-rate; \textbf{Late}: no-rate at $\geq$80\% revealed; \textbf{Retain}: \% of peak no-rate retained after local peaks; \textbf{Infl$_{60}$}: fraction of sharp ($\leq$60°) reversals. \emph{New Yorker} leads on all four while underperforming top LLMs on the rubric.}
\label{tab:combined-validation}
\end{table*}

Table~\ref{tab:combined-validation} and Figure~\ref{fig:norate-validation} report the results. On EQ-Bench, LLMs (81--84) comfortably outscore all human datasets (44--78), including the \emph{New Yorker} stories. EQ-Bench's LLM stories are produced in a single turn (with no system message, revision, or structural planning) from 32 genre-specific writing prompts, each 75--560 words specifying genre, character, setting, and stylistic constraints (e.g., a noir romance at Coney Island in 1937). \emph{New Yorker} stories, by contrast, are written by professional authors, typically drafted and revised over months, and selected through a competitive editorial process at a magazine that has published literary fiction since 1925. It is implausible that a single-pass LLM could genuinely surpass this level of craft; if taken at face value, it would mean the craft of literary fiction has been solved—a claim directly contradicted by various human validation results from recent NLP research in this area \citep{marco2024pron,chakrabarty2024art,tian2024narratives,marco2025reader,ismayilzada2025evaluating}. The more likely explanation is that the rubric misses something important, and EQ-Bench rewards dimensions where LLMs already excel (e.g., prose fluency, structural completeness, and instruction-following), while omitting narrative tension, the dimension where professional fiction most clearly surpasses LLM output.

Our \textit{100-Endings} metric provides confirmation of our central hypothesis: LLMs cannot hold tension. \textit{100-Endings} orders the four groups consistently with intuition: \textit{New Yorker} (0.765) exceeds EQ-Bench's top-10 LLMs (0.63) by $+$0.14, and both amateur corpora fall in between (Table~\ref{tab:combined-validation}). The gap is especially stark in late-stage unpredictability: \emph{New Yorker} stories maintain a late-stage no-rate of 0.607, nearly triple that of LLMs (0.215), which means that professional fiction preserves narrative tension even as it approaches its ending while LLM stories become predictable well before they conclude. Post-spike retention reinforces this: \emph{New Yorker} stories retain 52\% of local peaks on average, while LLMs retain only 23\%. Professional authors raise narrative questions and leave them partially open; LLMs resolve them almost immediately. We argue that this is the missing dimension in EQ-Bench—the sustained management of what the reader does and does not yet know.

Neither story length nor genre explains this ordering: Appendix~\ref{app:confounders} reports length-controlled regressions, per-genre means, and a matched-genre comparison.

\subsection{Human Evaluation}
\label{sec3.3}
While compelling, the recovery of a plausible corpus ordering is nonetheless indirect evidence. We run two studies with five expert annotators to further validate \textit{100-Endings} against human readers directly, one at the level of the whole story the other of the individual verdicts the metric is built from. Appendix~\ref{app:human-eval} gives the full study protocol, annotator instructions, and detailed results.

\paragraph{Does no-rate track story-level tension?}
Six generated stories are rated on a 1--5 scale for how well each retains narrative tension across its arc. Panel-mean ratings track both mean and late no-rate at Spearman $\rho = +0.657$, meaning that stories with higher no-rate are, in rank order, judged as more tension-rich by the panel of readers with high inter-rater agreement (Krippendorff's $\alpha = 0.93$, ICC(2,\,5) $= 0.92$, Kendall's $W = 0.66$, $p = 0.006$). Four of five annotators independently recover the metric's ordering of high/medium/low tension stories, and the panel agrees with no-rate on 11 of 15 pairwise comparisons and on all five of the widest no-rate gaps (the four disagreements all fall between stories the metric scores closely). 

\paragraph{Does the LLM judge match human verdicts?}
To mirror the LLM judge's evaluation settings, three of the annotators then label 600 generated endings drawn from the same six stories for their similarity to the actual continuation. The model matches the human-majority verdict on 68.3\% of generated endings ($\kappa = 0.362$), within 2\% of how often two independent humans agree with each other on the same rows (70.5\%, pooled $\kappa = 0.395$). The moderate human agreement is the ceiling for what any model judge can achieve on a task that is inherently subjective.

\section{Pipeline}
\label{sec4}

\paragraph{Story Idea Primer.}
The pipeline takes a story idea of any length as its sole input. These ``seed'' ideas can be user-supplied, drawn from existing benchmarks (e.g., the 32 genre-specific EQ-Bench prompts we use in Section~\ref{sec5}), or produced by a custom LLM idea-generation module combining genre profiles, tension-rich narrative tropes, or other mechanisms for combinatorial creativity (Appendix~\ref{app:input-idea}). Given an idea, the pipeline (Figure~\ref{fig:flow_chart}) produces a full story in three sequential steps, threading a shared context so that each step conditions on all prior outputs.

\begin{figure}[]
    \centering
    \includegraphics[width=0.95\linewidth]{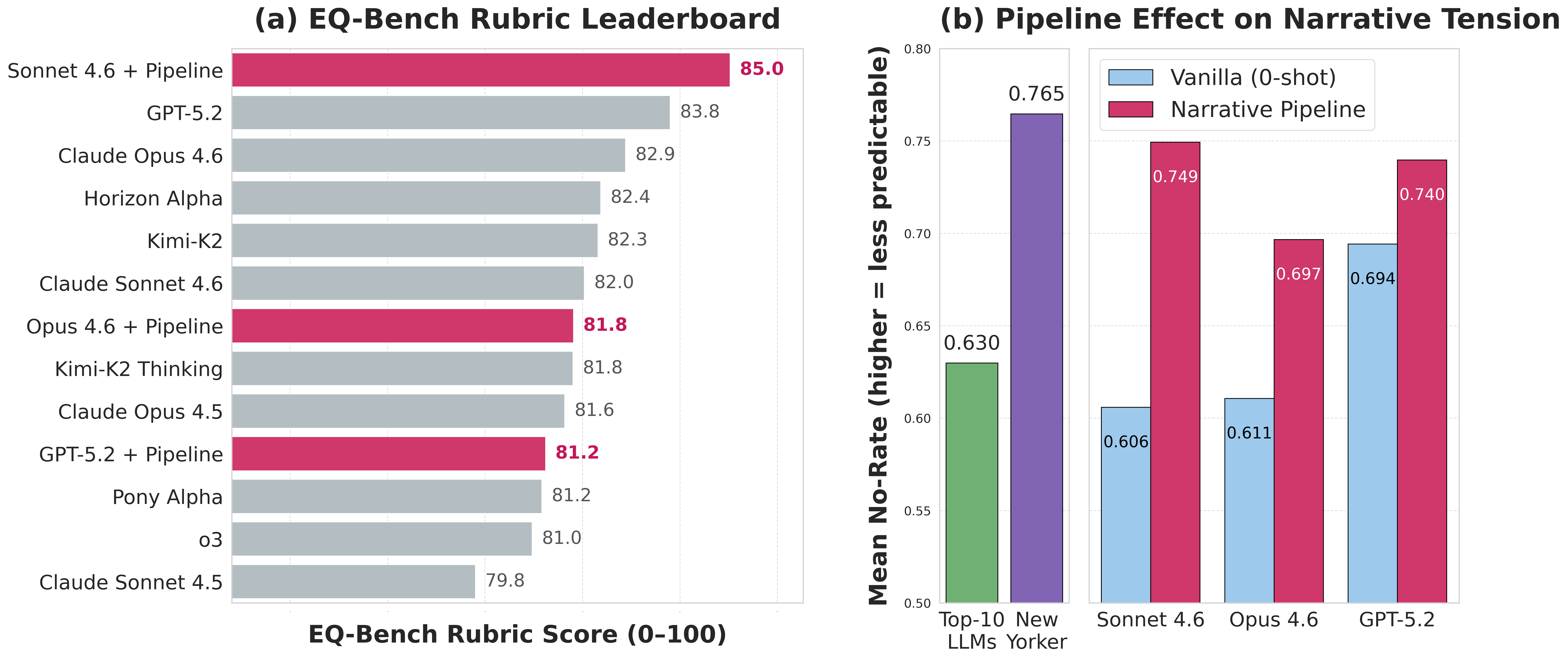}
    \caption{Effect of our narrative pipeline on rubric scores and tension metrics (32 EQ-Bench prompts). Left: EQ-Bench rubric leaderboard with our pipeline at \#1 (85.03). Right: \textit{$100$-Ending} on the same 32 prompts. The pipeline improves tension over vanilla models.}
    \label{fig:main-results}
\end{figure}

\paragraph{Step 0: Warmup Analysis.}
We randomly select a reference story either from our \emph{New Yorker} dataset or from a curated pool of canonical short stories; for the latter we supply only the title, leveraging the model's parametric knowledge. An LLM (Claude Sonnet, $T{=}0.7$) reads or recalls it and extracts 7 scene-level structural beats, each annotated with (1)~the concrete action, (2)~the specific narrative technique employed, and (3)~what made it memorable. This ``warmup link'' primes the model with the structural logic of high-quality literary fiction before it encounters the target story idea.

\paragraph{Step 1: Beat Adaptation (the Thick Intermediate Layer, or Structured Beat Scaffold).}
The extracted beat sheet is adapted to the target story idea. The LLM produces a detailed intermediate plan, a structural ``to-do list'' specifying for each beat the narrative function, what information is revealed or withheld, the tension mechanism in play, and how stakes escalate. This representation is deliberately \emph{thick}, in contrast to the thin, one-line outline a typical planning prompt returns: it carries enough structural information to constrain the generated story away from default narrative templates while leaving stylistic-level decisions to the next step. When the warmup step is ablated, beats are generated from scratch, prompted with the story idea alone.

\paragraph{Step 2: Story Generation.}
The LLM generates the final story conditioned on the story idea and plan from Step~1, instructed to follow the structural to-do list closely with a system-prompt emphasis on sustaining tension. When both planning steps are ablated, the pipeline reduces to zero-shot (vanilla) generation, providing a controlled baseline.

\begin{table}[t]
\centering
\small
\setlength{\tabcolsep}{4pt}
\caption{Pipeline effect on rubric quality and tension metrics (32 EQ-Bench prompts). Values are point estimates $\pm$ half-width of a 95\% percentile bootstrap interval over stories (exact bounds in Appendix~\ref{app:significance}), from iteration~1; the 4 Sonnet~4.6 repeats serve only the stability check in Appendix~\ref{app:stability}. $^\dagger$ marks a gain significant under a separate \emph{paired} bootstrap over the 32 shared prompts, which is more powerful than the $\pm$ intervals shown.}
\label{tab:pipeline-100ending}
\begin{tabular}{lcccccc}
\toprule
\textbf{Source} & $n$ & \textbf{EQ-Bench} & \textbf{Mean NR} & \textbf{Late NR} & \textbf{Retain} & \textbf{IR$_{60}$} \\
\midrule
Sonnet 4.6                 & 32 & 82.0 \tiny$\pm$1.7 & 0.606 \tiny$\pm$.063 & 0.137 \tiny$\pm$.042 & 14.6 \tiny$\pm$3.9 & 0.073 \tiny$\pm$.021 \\
\quad \textbf{+ Pipeline}  & 32 & \textbf{85.0} \tiny$\pm$2.3 & \textbf{0.749}$^\dagger$ \tiny$\pm$.044 & \textbf{0.340}$^\dagger$ \tiny$\pm$.068 & \textbf{34.1}$^\dagger$ \tiny$\pm$4.8 & \textbf{0.135}$^\dagger$ \tiny$\pm$.024 \\
\addlinespace
Opus 4.6                   & 32 & 82.9 \tiny$\pm$1.2 & 0.611 \tiny$\pm$.061 & 0.185 \tiny$\pm$.047 & 15.9 \tiny$\pm$3.1 & 0.090 \tiny$\pm$.022 \\
\quad \textbf{+ Pipeline}  & 32 & 81.8 \tiny$\pm$2.7 & \textbf{0.697}$^\dagger$ \tiny$\pm$.051 & \textbf{0.296}$^\dagger$ \tiny$\pm$.064 & \textbf{32.8}$^\dagger$ \tiny$\pm$5.3 & \textbf{0.142}$^\dagger$ \tiny$\pm$.031 \\
\addlinespace
GPT-5.2                    & 32 & 83.8 \tiny$\pm$1.5 & 0.694 \tiny$\pm$.067 & 0.329 \tiny$\pm$.072 & 36.1 \tiny$\pm$5.2 & 0.131 \tiny$\pm$.022 \\
\quad \textbf{+ Pipeline}  & 32 & 81.2 \tiny$\pm$2.3 & 0.740 \tiny$\pm$.059 & \textbf{0.443}$^\dagger$ \tiny$\pm$.069 & \textbf{43.6}$^\dagger$ \tiny$\pm$4.9 & 0.141 \tiny$\pm$.029 \\
\midrule
Top-10 LLMs (0-shot)       & 320 & 81.9 \tiny$\pm$0.6 & 0.630 \tiny$\pm$.021 & 0.215 \tiny$\pm$.018 & 23.3 \tiny$\pm$1.6 & 0.106 \tiny$\pm$.008 \\
\emph{New Yorker}          & 261 & 78.6 \tiny$\pm$2.8 & 0.765 \tiny$\pm$.019 & 0.607 \tiny$\pm$.027 & 51.7 \tiny$\pm$2.2 & 0.200 \tiny$\pm$.011 \\
\bottomrule
\end{tabular}
\end{table}

\section{Results}
\label{sec5}

We investigate whether structured constraints can improve long-range narrative tension while maintaining or even enhancing other metrics of fiction quality. Despite the limitations of the EQ-Bench rubric demonstrated in Section~\ref{sec3}, it remains one of the most widely used standards and captures dimensions of story quality orthogonal to tension. We apply our full three-step pipeline (Section~\ref{sec4}) to three frontier models (Claude Sonnet~4.6, Claude Opus~4.6, and GPT-5.2) across the standard 32 EQ-Bench prompts, and evaluate the outputs on both the EQ-Bench rubric and \textit{100-Endings}.

Table~\ref{tab:pipeline-100ending} reports the results with 95\% bootstrap intervals. Late-stage no-rate and post-spike retention improve significantly for all three models, lifting each well above the top-10 LLM baseline (0.215) and toward \emph{New Yorker} levels (0.607), where LLM writing most clearly lags professional fiction. Mean no-rate improves significantly for Sonnet~4.6 and Opus~4.6, though not for GPT-5.2 ($n{=}32$), whose zero-shot outputs already score highest of the three. Gains are largest for Sonnet~4.6: mean no-rate rises from 0.606 to 0.749, late-stage no-rate more than doubles, retention rises from 14.6\% to 34.1\%, and the EQ-Bench score also improves, placing it \#1 on the leaderboard (Figure~\ref{fig:main-results}). No EQ-Bench difference is significant in either direction, including the nominal $-$1.1 and $-$2.6 point drops for Opus~4.6 and GPT-5.2, so the pipeline's tension gains do not come at the cost of what the rubric does measure.

\section{Discussion: What \textit{100-Endings} Measures}
\label{sec6}

\begin{figure}[t]
    \centering
    \includegraphics[width=0.58\textwidth]{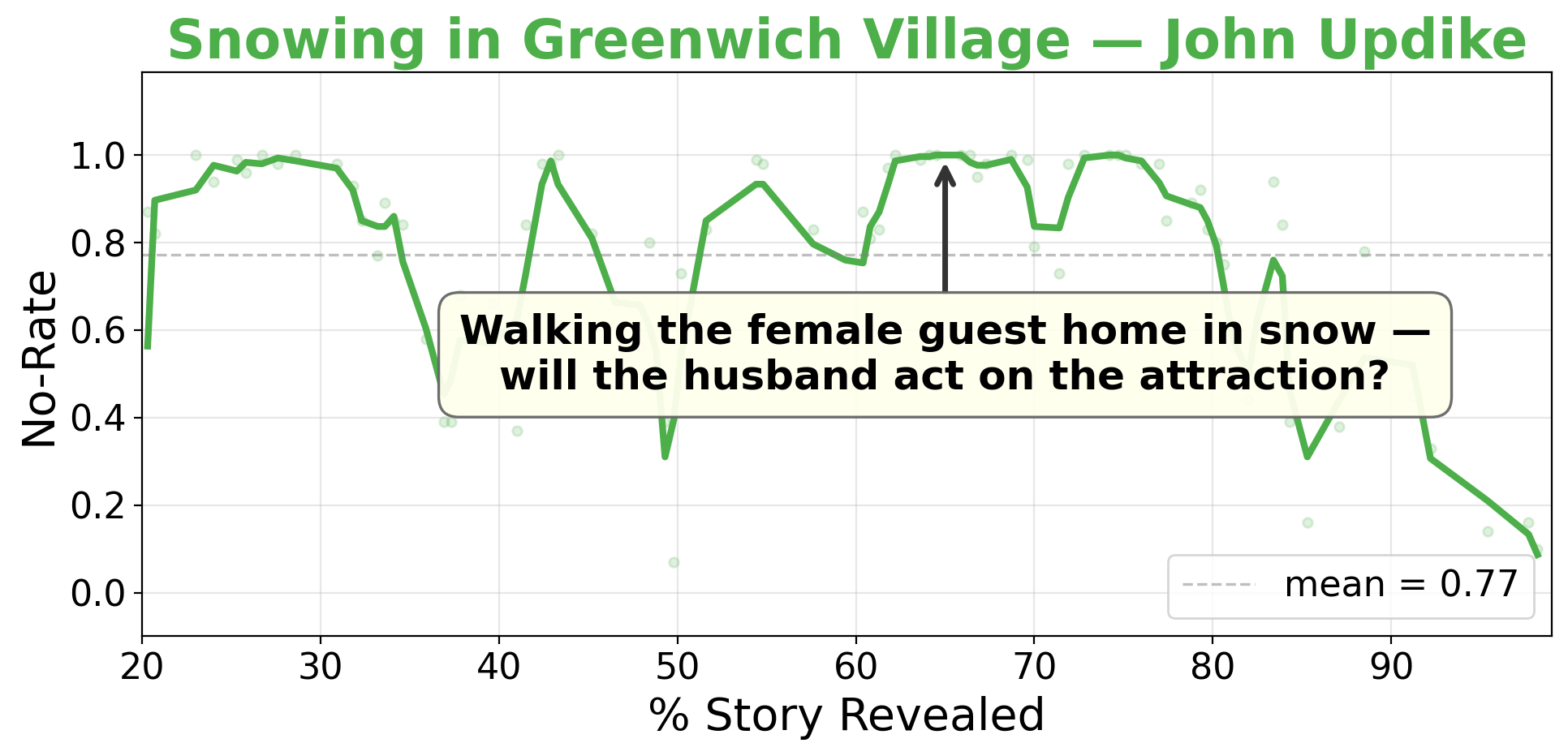}
    \caption{No-rate curve for Updike's ``Snowing in Greenwich Village.'' Tension rises once the husband walks the guest home alone (${\sim}60\%$ story revealed) and stays elevated through the ambiguous ending.}
    \label{fig:updike-norate}
\end{figure}

\begin{figure}[t]
    \centering
    \includegraphics[width=0.70\textwidth]{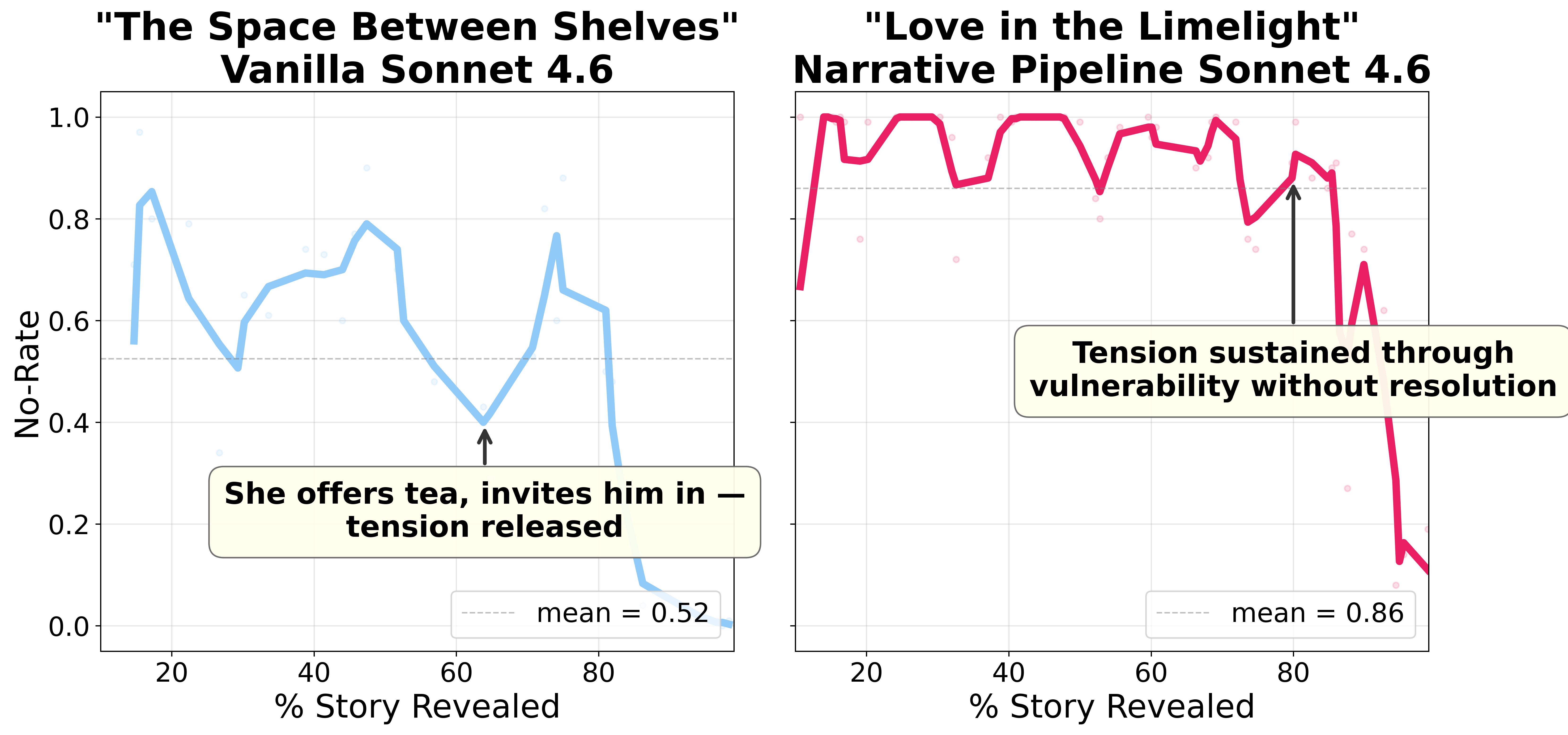}
    \caption{No-rate curves for two stories generated by Sonnet~4.6 from the same prompt. \textbf{Left:} Zero-shot. \textbf{Right:} Narrative pipeline (warmup + planning). The zero-shot curve collapses early, while the pipeline sustains high unpredictability. See Appendix~\ref{app:case_study_stories} for full story texts.}
    \label{fig:bookshop-contrast}
\end{figure}

\textit{100-Endings} is a measure of predictability, which we argue is a necessary (but not sufficient) component of tension. A story can be unpredictable and incoherent, since tension without stakes, investment, or logic is simply chaos. The metric should therefore be used alongside mainstream benchmarks of generic writing quality; separating genuinely compelling high-tension stories from merely erratic ones remains an open problem. Yet, the metric's ability to distinguish professional fiction from LLM outputs (unlike EQ-Bench) suggests it captures an essential element of narrative craft: the deliberate management of reader expectations through information asymmetry, delayed resolution, and structural surprise.

While our four statistics (Section~\ref{sec3}) primarily evaluate \emph{plot-level} predictability, the generated endings offer a rich artifact for further analysis. Future work could develop metrics analyzing the \emph{style} of these endings to capture sub-event forms of tension, such as thematic irony and psychological interiority.

The metric's reliance on LLMs as generator raises a circularity concern: are we measuring the properties of the stories, or the idiosyncratic preferences of the judge? We mitigate this by strictly separating the generator and evaluator models (employing \textsc{Qwen3} exclusively for evaluation) and by validating that the metric correctly orders human corpora of known quality (details see Appendix~\ref{app:datasets}).

\paragraph{The Gap to Professional Fiction}

Our best pipeline configuration achieves a mean no-rate of $0.749$ on the EQ-Bench prompts, closing most of the gap between vanilla Sonnet~4.6 ($0.606$) and \textit{New Yorker} fiction ($0.765$). On late-stage no-rate, where the deficit is widest, it closes roughly half ($0.137$ to $0.340$, against $0.607$). The remaining gap likely reflects aspects of professional craft that our pipeline does not yet address: tacit knowledge based on cultural contexts, subversion of genre through deep familiarity with their conventions, and the kind of sentence-level surprise that emerges from a distinctive authorial voice rather than structural planning.

\paragraph{What is the Shape of Stories With Tension?}
The central finding of this work is that zero-shot LLMs tend to collapse narrative tension through premature resolution, a shortcoming that a thick, structured generation pipeline can mitigate. To illustrate the problem and the remedy, we examine the tension curves of three romance stories that share a single source of suspense: the unspoken chemistry between a man and a woman alone together. One is from our \textit{New Yorker} dataset (Figure~\ref{fig:updike-norate}), while the other two are generated by \textsc{Sonnet~4.6} from identical prompts with versus without our pipeline (Figure~\ref{fig:bookshop-contrast}). Because the genre, subject, and emotional register are held constant, the resulting differences in the no-rate curves isolate how effectively each story manages the gap between possibility and resolution.

\section{Conclusion and Future Work}
Zero-shot LLMs struggle to hold narrative tension, resolving uncertainty prematurely rather than sustaining the gap between reader knowledge and narrative withholding. Because tension is downstream of hidden information (secrets, delayed revelations, chilling endings), this tendency to prematurely reveal produces stories that are polished but flat. The \textit{100-Endings} metric makes this deficit measurable, operationalizing tension as unpredictability of endings at the sentence level. Our results reveal that stories generated by top LLMs exhibit the lowest tension of any evaluated corpus, even when rubric-based judges rate them above professional fiction. However, our structured generation pipeline demonstrates that this deficit is not immutable: literary theory-driven constraints can significantly improve long-range tension. While tension is a vital aspect of good storytelling, it is only one element of the broader space of key narrative attributes required to build true narrative machines.

This paper provides a framework for quantifying and improving narrative tension, and represents an initial step toward a comprehensive computational narratology \citep{piper2021narrative} that opens up several interesting directions. Our immediate subsequent work will expand on various components of our pipeline, particularly the generation of synthetic story primers at scale. We also plan to conduct systematic ablations across every stage of the pipeline to isolate which structural constraints contribute most to tension—specifically, whether the key factor is the thickness of intermediate steps or the combinatorial novelty of the constraints themselves. Aside from measuring tension, the \textit{100-Endings} framework also produces a rich object (100 candidate endings at every sentence boundary) from which further metrics could be derived, particularly literary theory-driven measures of stylistic or thematic aspects of quality that operate below the level of overall tension. 

Beyond the pipeline, tension is only one dimension of narrative craft. Good stories do not emerge from a vacuum, and our future work aims to explore the tacit knowledge and contextual frameworks behind narrative artifacts like tension—if multi-agent architectures can simulate the editorial processes, social dynamics, and cultural ecosystems from which high-quality literature originates. Modeling these environments may prove essential for closing the remaining gap to professional fiction.

\section*{Ethics Statement}
Our research involves no human subjects and utilizes publicly available or archived literary datasets. To ensure ethical compliance, the \textit{New Yorker} data is stored in a secure repository at our university; its use in this study constitutes fair use, as it is employed strictly for non-disseminating, non-commercial research. Other amateur corpora, such as StoryStar, were sourced from public-facing websites intended for open readership.

\paragraph{Human Annotation.}
The annotators in our human evaluation (Section~\ref{sec3.3}, Appendix~\ref{app:human-eval}) are fairly compensated at \$15USD/hour. They were informed the general purpose of the study, read and rated model-generated stories only, and are not identified in this paper or any released data.

\paragraph{Impact on Human Creators.}
Utilizing computational tools to operationalize narrative tension allows us to map the formal contours of what is often felt as a nebulous spark of creativity. We hope this work has a positive impact on every person who takes satisfaction in a story finding its shape.  Our objective is to demystify these structures to empower, rather than displace, the author. As stated in our humanistic mission statement, we do not intend for these systems to replace human storytellers; instead, we hope that by improving machine performance on these tasks, we create tools that augment the human imagination.

\paragraph{Disclosure of LLM Use.}
Language models are both the object of study and the instrument of measurement in this work: the corpora we evaluate include model-generated stories, and the \textit{100-Endings} metric is itself implemented with an LLM generator and an LLM judge, as described in Section~\ref{sec3} and Appendix~\ref{app:metric}. Beyond that, we used LLMs to assist with paper editing and formatting BibTeX entries.

\section*{Acknowledgments}
This work was supported, in part, by the Schmidt Sciences Humanities and AI Virtual Institute (HAVI) and the “Humanistic AI” grant from the Neubauer Collegium for Culture and Society, University of Chicago. We thank the Textual Optics Lab at the University of Chicago for providing access to \textit{The New Yorker} corpus.

\bibliography{colm2026_conference}
\bibliographystyle{colm2026_conference}

\newpage
\appendix

\section*{Humanistic Mission Statement\footnote{Following \citet{sui2025kristeva}, we include this statement to ensure that our AI research faithfully adheres to the standards and values of humanities scholarship.}}
This work aims to broaden the intersection of AI research and the humanities, specifically literary theory. As LLMs become more deeply embedded in society, NLP research is growing increasingly interdisciplinary, encountering nuanced problem spaces that prove to be recalcitrant to the expertise and methods of computer science alone. We argue that it is vital to bridge NLP research with existing domain knowledge in relevant disciplines, such as applying narrative theory to computational narrative understanding \citep{piper2021narrative}. Just as AI research has long embraced physics-informed machine learning to ground models in the natural world, we advocate for the development of literary-theory-informed models for computational creativity. We hope to encourage more collaborative work between NLP researchers and humanities scholars, ensuring that future models are grounded in the rigorous theoretical frameworks of the domains they attempt to emulate.

We strongly oppose the use of AI research to disrupt the craft of storytelling or even displace human writers—for instance, by building LLM systems that claim to write "better" stories in some shape or form. Our goal is to develop computational approaches that can study narrative and its affective powers at scale. Likewise, our \textit{100-Endings} metric is not designed to reduce the quality of fiction to an algorithmic monolith, but to introduce a reliable and scalable yardstick to help us better understand the complex interplay between various mechanisms of storytelling.

\paragraph{Scope.}
We validate \textit{100-Endings} on short fiction ($\leq$5{,}000 words). Beyond that range both the generator's context window and the construct itself degrade: forecasting a novel's ending from its opening chapters approaches the unconditional base rate, and we would expect no-rate to saturate. We therefore make no claims about long-form narrative, and leave a chapter-level variant of the metric to future work.

\newpage

\section{Datasets}
\label{app:datasets}

We evaluate on four story corpora spanning professional fiction, amateur writing, and LLM-generated text. Table~\ref{tab:corpora-summary} summarizes the key statistics.

\begin{table}[h]
\centering
\small
\caption{Summary of evaluation corpora. ``100-Ending $n$'' is the number of stories used for the 100-ending metric (500--5{,}000 word range); ``EQ-Bench $n$'' is the number used for EQ-Bench rubric evaluation (2{,}000--3{,}000 word range).}
\label{tab:corpora-summary}
\begin{tabular}{llccc}
\toprule
\textbf{Corpus} & \textbf{Quality tier} & \textbf{Total stories} & \textbf{100-Ending $n$} & \textbf{EQ-Bench $n$} \\
\midrule
New Yorker         & Professional & 261  & 261 & 53 \\
Tell Me a Story    & Amateur (curated) & 230  & 100 & 41 \\
StoryStar          & Amateur (scraped) & 100 & 100 & 31 \\
EQ-Bench LLMs      & LLM-generated & 32 $\times$ $M$ & 320 & 320 \\
\bottomrule
\end{tabular}
\end{table}

\paragraph{New Yorker ($n = 261$).} We utilize a corpus of short fiction published in \textit{The New Yorker} between 1945 and 2019, widely regarded as a gold standard for contemporary literary fiction. To ensure ethical compliance, the data is stored in a secure repository at our university; its use in this study constitutes fair use, as it is employed strictly for non-disseminating, non-commercial research. From the total dataset, we select 261 stories under 5,000 words for the \textit{100-Ending} analysis. For the EQ-Bench rubric evaluation, which requires a narrow word-count band to control for length bias (Section~\ref{app:experimental-details}), 53 stories fall within the 2,000--3,000 word range.

\paragraph{Tell Me a Story ($n_{\text{sample}} = 100$).} Tell Me a Story (TMAS) is a publicly available dataset of short stories written by amateur authors in a guided creative-writing setting \citep{huot2025agents}. The full corpus contains 230 stories. We randomly sample 100 stories (500--5{,}000 word range) for the 100-ending evaluation. For the EQ-Bench rubric evaluation, 41 stories fall within the 2{,}000--3{,}000 word band.

\paragraph{StoryStar ($n = 100$).} StoryStar is a corpus of 100 amateur short stories scraped from \texttt{storystar.com}, a public creative-writing platform. Stories are sampled to fall within the 500--5{,}000 word range. Unlike TMAS, StoryStar stories receive no editorial curation, making this corpus a lower-quality amateur baseline.

\paragraph{EQ-Bench LLMs ($n = 320$).} We include zero-shot LLM-generated stories from the top-performing models on the EQ-Bench Creative Writing leaderboard~\citep{creative-writing-bench-v3}. Each model generates one story per each of the 32 standard EQ-Bench prompts, sourced directly from the benchmark's public repository.\footnote{\url{https://github.com/EQ-bench/creative-writing-bench}} We use outputs from the top-10 ranked models at the time of evaluation. After filtering to the 500--5{,}000 word range, this yields 320 stories. These serve as the baseline: stories that score highly on existing rubric-based metrics but, as we show, lack narrative tension.

\section{The EQ Bench Creative Writing Baseline}
\label{app:experimental-details}

\begin{figure}[t]
    \centering
    \includegraphics[width=0.95\columnwidth]{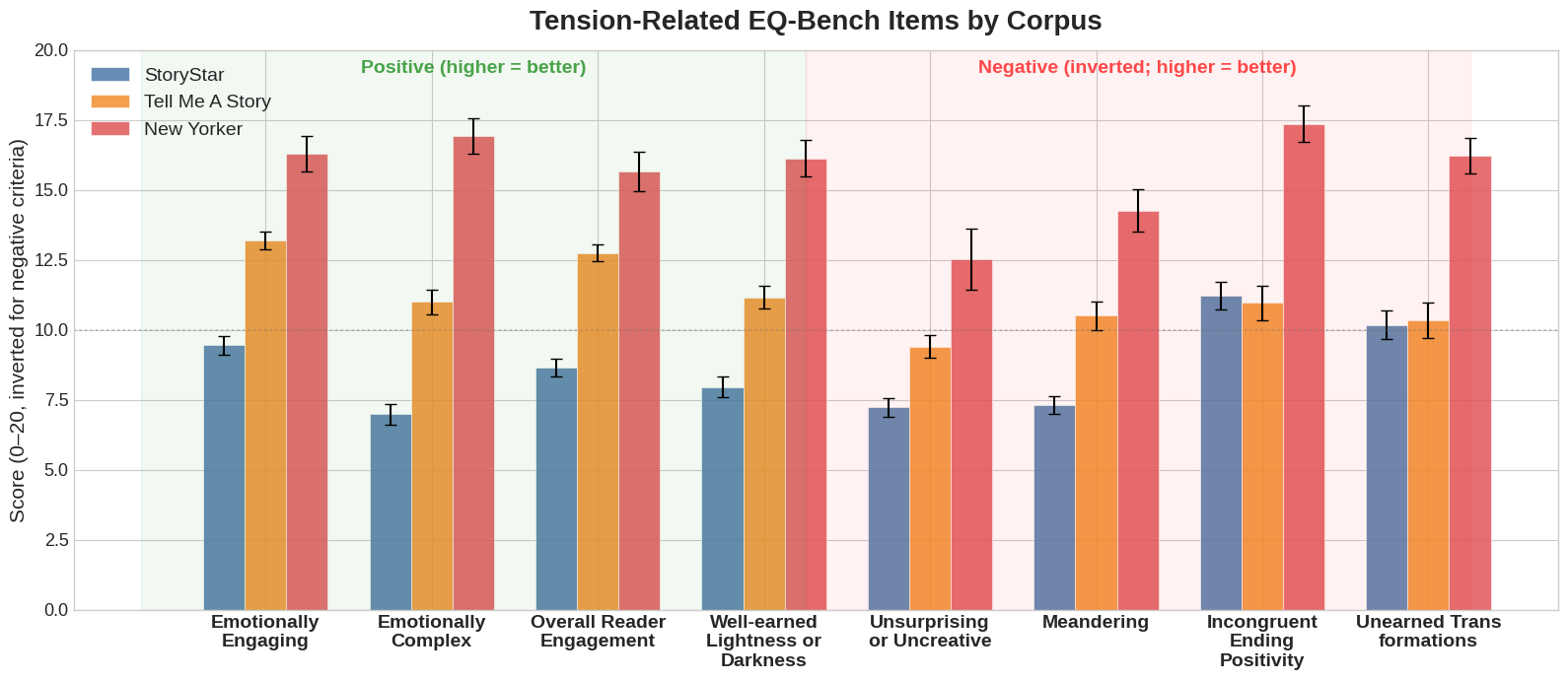}
    \caption{Tension-related EQ-Bench sub-criteria across three human-written
    corpora (higher~$=$~better; negative criteria are inverted). The rubric
    captures some tension-related signal but evaluates it as a holistic
    \emph{judgment}—not through the operational lens of how a story
    manages information and prediction over time.}
    \label{fig:tension-criteria}
\end{figure}

\paragraph{EQ-Bench has tension-adjacent criteria, but they are insufficient.}
The rubric is not entirely blind to tension. Among its 22 criteria are several that relate to engagement and surprise: \emph{Emotionally Engaging}, \emph{Emotionally Complex}, \emph{Overall Reader Engagement}, \emph{Well-earned Lightness/Darkness}, and (inverted) \emph{Unsurprising / Uncreative}, \emph{Meandering}, \emph{Incongruent Ending Positivity}, and \emph{Unearned Transformations}. Figure~\ref{fig:tension-criteria} shows that these criteria do differentiate the three human corpora in the expected direction: \emph{New Yorker} fiction substantially outperforms amateur writing on every tension-related sub-score. Obviously, these criteria can indeed distinguish good writing from bad writing. 

Yet these criteria evaluate tension as a \emph{holistic rubric judgment}: a judge reads the complete story and rates how ``surprising'' or ``engaging'' it felt. This approach has two structural limitations. First, rubric criteria like ``Unsurprising / Uncreative'' could conflate multiple constructs—an inventive premise can score well even if the plot unfolds predictably. Second, a single post-hoc score cannot capture \emph{where} and \emph{how} tension operates across the story arc: whether information is withheld, whether resolution is delayed, or whether the narrative resists prediction at each point in its progression. As a result, EQ Bench cannot distinguish great stories from ones that just have good prose. As our formal definitions in Section~\ref{sec2} make clear, measuring tension requires a \textbf{position-level} evaluation that tracks predictability across the entire arc.

\paragraph{Length-controlled evaluation.}
LLM judges exhibit a strong length bias in pairwise and rubric-based evaluation: longer outputs systematically receive higher scores~\citep{creative-writing-bench-v3}. EQ-Bench mitigates this by truncating outputs at 4{,}000 characters, placing all models on equal footing. We adopt an analogous strategy for corpus evaluation: we restrict the rubric assessment to stories in a narrow word-count band (2{,}000--3{,}000 words), ensuring that differences in rubric scores reflect writing quality rather than length. This yields 31 StoryStar stories, 41 TMAS stories, and 53 New Yorker stories for the human corpora.

For LLM-generated stories, most zero-shot outputs from the top-10 EQ-Bench models are substantially shorter than 2{,}000 words: across 320 stories (10 models $\times$ 32 prompts), the median length is 1{,}102 words, with only 14\% falling in the 2{,}000--3{,}000 word range. GPT-5.2 is the exception, with 23 of its 32 stories in the target range. We therefore report two LLM entries: GPT-5.2 restricted to 2{,}000--3{,}000 words (the length-controlled comparison), and the pooled top-10 at all lengths. Note that the length restriction \emph{disadvantages} LLMs: EQ-Bench's own documentation notes that shorter outputs receive lower rubric scores. The fact that even at all lengths—where LLMs are penalized by their own brevity—they still outscore all human corpora makes the dissociation with our 100-ending metric all the more striking.

\paragraph{Full rubric results.}
Table~\ref{tab:corpus-rubric} reports the EQ-Bench rubric scores across all evaluated groups. All evaluations use Claude Sonnet~4 as judge, following the standard EQ-Bench protocol~\citep{creative-writing-bench-v3}.

\begin{table}[t]
\centering
\small
\caption{EQ-Bench Creative Writing rubric scores (0--100 scale). \textbf{Overall}: mean across all 22 criteria. \textbf{Tension$_8$}: mean across the 8 tension-related criteria only (\emph{Emotionally Engaging}, \emph{Emotionally Complex}, \emph{Overall Reader Engagement}, \emph{Well-earned Lightness/Darkness}, and inverted \emph{Unsurprising/Uncreative}, \emph{Meandering}, \emph{Incongruent Ending Positivity}, \emph{Unearned Transformations}). Human corpora restricted to 2{,}000--3{,}000 words; GPT-5.2 similarly restricted; pooled top-10 includes all lengths. Judge: Claude Sonnet~4.}
\label{tab:corpus-rubric}
\begin{tabular}{lcccc}
\toprule
\textbf{Source} & \textbf{Overall} & \textbf{Tension$_8$} & $\pm$\textbf{95\% CI} & $n$ \\
\midrule
GPT-5.2 (2--3k words)             & 84.28 & 83.86 & 1.81 & 23 \\
Top-10 LLMs (all lengths)         & 81.88 & 80.80 & 0.55 & 320 \\
\midrule
\emph{New Yorker} (professional)  & 78.61 & 78.40 & 4.33 & 53 \\
Tell Me a Story (amateur)         & 60.89 & 55.85 & 3.05 & 41 \\
StoryStar (amateur)               & 45.81 & 44.17 & 4.23 & 31 \\
\bottomrule
\end{tabular}
\end{table}

GPT-5.2, even when restricted to the same 2{,}000--3{,}000 word range as the human corpora, scores 84.28 overall—comfortably above \emph{New Yorker} fiction (78.61). The pooled top-10 at all lengths scores 81.88, still above professional fiction despite the length disadvantage. Notably, even on the 8 tension-related sub-criteria alone, LLMs outscore \emph{New Yorker}: GPT-5.2 scores 83.86 vs.\ 78.40 for professional fiction. This means the rubric's tension-adjacent criteria—which evaluate tension as a holistic post-hoc judgment—do not capture the same signal as our position-level 100-ending metric, which correctly reverses the ordering. Among human corpora, \emph{New Yorker} leads by a wide margin over Tell Me a Story (60.89 overall, 55.85 on tension) and StoryStar (45.81 overall, 44.17 on tension).

\section{The 100-Ending Metric}
\label{app:metric}

This section formalizes the 100-ending metric and provides complete implementation details, including exact prompt templates and all hyperparameters.

\subsection{Overview}
\label{app:metric-overview}

The 100-ending metric measures \textbf{narrative predictability} at the sentence level. The intuition is simple: at each point in a story, we ask a language model to predict how the story ends, repeating this 100 times, then check how many predictions match what actually happens. If the true ending is easy to predict, most predictions will match; if it is surprising, few will.

Crucially, the model is asked to \emph{describe} how the story ends—a plot-level prediction—rather than to write creative prose. This frames the task as forecasting rather than generation, measuring whether the model can anticipate the narrative trajectory.

Formally, let $S = (s_1, s_2, \ldots, s_N)$ be a story decomposed into $N$ sentences. At each sentence boundary $i \in \{1, \ldots, N{-}1\}$, we:

\begin{enumerate}[leftmargin=2em]
    \item \textbf{Predict} 100 independent ending predictions $\hat{e}_1^{(i)}, \ldots, \hat{e}_{100}^{(i)}$ conditioned on the prefix $(s_1, \ldots, s_i)$, using a generation model $\mathcal{G}$.
    \item \textbf{Judge} each prediction against the true remainder $(s_{i+1}, \ldots, s_N)$ using a judge model $\mathcal{J}$, which returns a binary verdict $j_k^{(i)} \in \{0, 1\}$ indicating whether $\hat{e}_k^{(i)}$ captures the ``general direction'' of the true continuation.
    \item \textbf{Aggregate} into a position-level \textbf{no-rate}:
    \begin{equation}
        \text{no-rate}(i) = 1 - \frac{1}{100} \sum_{k=1}^{100} j_k^{(i)}
    \end{equation}
\end{enumerate}

\noindent A no-rate of 0.85 means 85 of 100 predicted endings fail to match what actually happens—the story is highly unpredictable at that point. The \textbf{story-level no-rate} is the mean across all evaluated positions:
\begin{equation}
    \overline{\text{no-rate}} = \frac{1}{|\mathcal{P}|} \sum_{i \in \mathcal{P}} \text{no-rate}(i)
\end{equation}
where $\mathcal{P}$ is the set of retained positions (see Section~\ref{app:position-selection}).

\subsection{Sentence Splitting}
\label{app:sentence-splitting}

Stories are split into sentences using a rule-based boundary detector. The algorithm identifies boundaries where terminal punctuation (\texttt{.}, \texttt{!}, \texttt{?}), optionally followed by closing quotation marks or parentheses, is followed by whitespace and then an uppercase letter or opening punctuation. The regex pattern is:

\begin{center}
\small
\verb|[.!?][""\u2019)]*\s+(?=[A-Z\u201c"(\[])|
\end{center}

\noindent This heuristic correctly handles most English prose, including quoted dialogue and parenthetical asides. Both the generation and judge pipelines use an identical implementation of this splitter.

\subsection{Position Selection}
\label{app:position-selection}

Not all sentence boundaries are evaluated. We apply two filters:

\begin{enumerate}[leftmargin=2em]
    \item \textbf{Minimum context}: Position $i$ is skipped if the preceding sentence $s_{i-1}$ contains fewer than 10 words (\texttt{MIN\_SENT\_WORDS\,=\,10}). This excludes positions following very short sentences (e.g., single-line dialogue tags) that provide minimal narrative context.
    \item \textbf{Token percentage window}: We compute the fraction of story tokens revealed at each position using the tokenizer of the generation model. Only positions where 10\%--99\% of tokens have been revealed are retained. The lower bound excludes positions with insufficient context; the upper bound excludes trivial positions near the final sentence.
\end{enumerate}

\noindent Stories exceeding 5{,}000 words are excluded entirely to ensure the full context fits within the generation model's context window at every position.

\subsection{Stage 1: Ending Generation}
\label{app:ending-gen}

\paragraph{Model.} We use \textbf{Qwen3-32B} as the generation model $\mathcal{G}$, served via vLLM with offline batch inference on a single NVIDIA H200 GPU. The vLLM engine is configured with \texttt{dtype=bfloat16}, \texttt{max\_model\_len=8192}, \texttt{enable\_prefix\_caching=True}, and \texttt{gpu\_memory\_utilization=0.92}. The model's chain-of-thought mode is disabled (\texttt{enable\_thinking=False}).

\paragraph{Prompt template.} At each retained position $i$, we construct a chat-formatted prompt with a system message and user message:

\begin{quote}
\small
\textbf{System:}\\
\texttt{You are a fiction writer. Write only story text.}

\vspace{0.5em}
\textbf{User:}\\
\texttt{In 2--3 sentences, describe how you think this story ends. No commentary, no explanation, no preamble.}

\texttt{STORY SO FAR:}\\
\texttt{\{$s_1 \;\| \; s_2 \;\| \; \cdots \;\| \; s_i$\}}
\end{quote}

\noindent where $\|$ denotes string concatenation with a space separator. The prompt asks the model to \emph{describe} the ending rather than write it as prose. This frames the task as plot-level forecasting: the model predicts what events occur, not how they would be narrated.

\begin{figure}[p] %
    \centering
    \vspace{-0.5cm}
    \vfill
    \includegraphics[width=0.85\columnwidth]{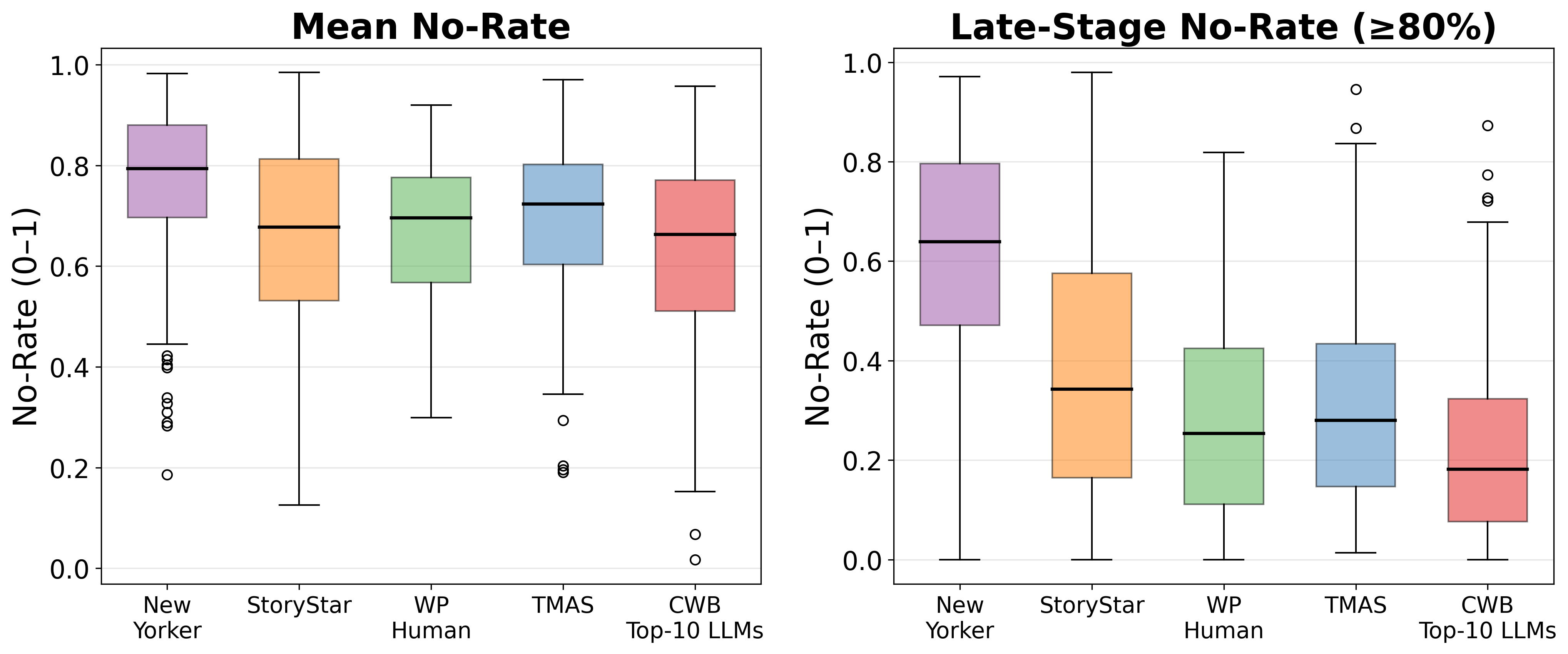}
    \caption{Overall (left) and late-stage ($\geq$80\%, right) mean no-rate by corpus. \emph{New Yorker} fiction is the least predictable on both measures. Top-10 LLMs score lowest, below all human corpora including amateurs, despite outscoring \emph{New Yorker} on EQ-Bench (Table~\ref{tab:corpus-rubric}).}
    \label{fig:norate-validation}

    \vspace{0.4cm} %

    \includegraphics[width=\columnwidth]{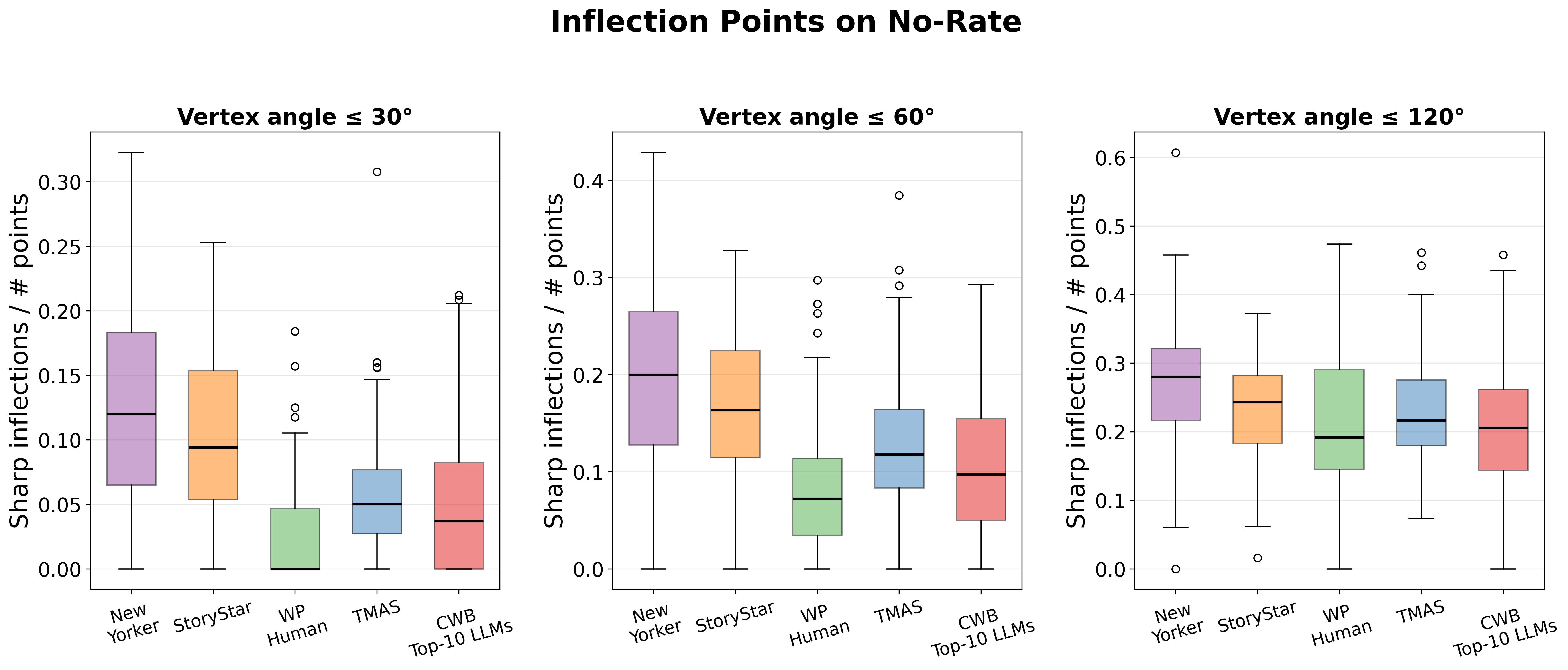}
    \caption{Inflection rate by corpus at three vertex-angle thresholds ($30°$, $60°$, $120°$). \emph{New Yorker} fiction has the highest inflection rate at all thresholds, indicating more frequent sharp reversals in the no-rate curve. The gap narrows at looser thresholds as all corpora accumulate more qualifying reversals.}
    \label{fig:inflection-by-corpus}

    \vspace{0.4cm}

    \includegraphics[width=0.45\columnwidth]{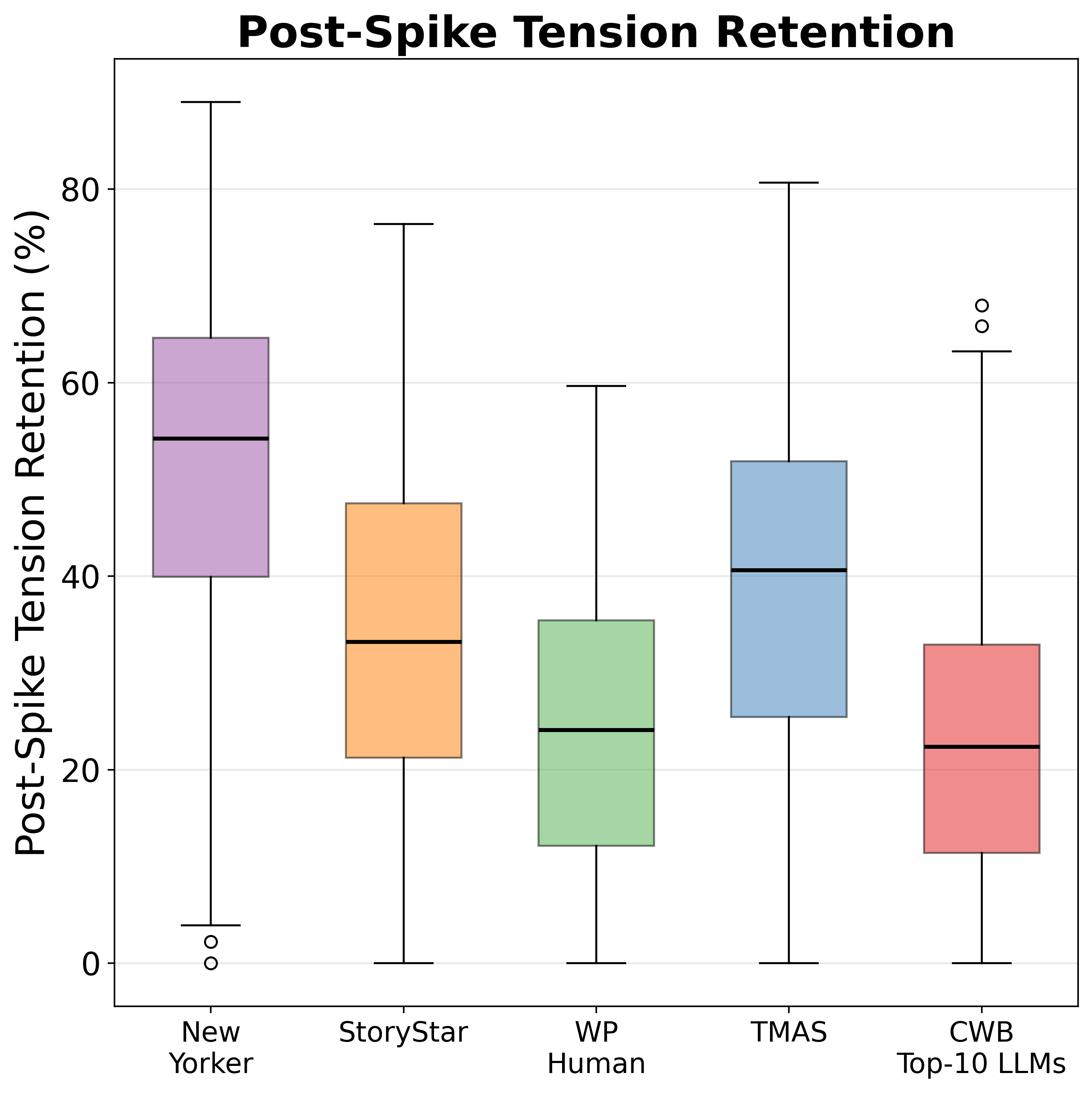}
    \caption{Post-spike tension retention (\%) by corpus. After a local peak in the no-rate curve, what fraction of the peak value is retained at the subsequent minimum? \emph{New Yorker} stories retain the most tension after peaks (median 54\%), while LLM stories resolve peaks almost completely (median 22\%).}
    \label{fig:retention-by-corpus}
    \vfill

\end{figure}

\paragraph{Sampling parameters.}
We generate $K = 100$ independent ending predictions per position using vLLM's \texttt{n=100} parameter (all 100 samples drawn in a single batched call per position). The sampling configuration is:

\begin{table}[h]
\centering
\small
\caption{Ending generation sampling parameters.}
\label{tab:gen-params}
\begin{tabular}{lcp{7.5cm}}
\toprule
\textbf{Parameter} & \textbf{Value} & \textbf{Rationale} \\
\midrule
Temperature & 1.2 & Elevated to produce diverse predictions that probe the full space of plausible endings. Lower temperatures collapse the prediction distribution, making it impossible to distinguish ``predictable because all readers would agree'' from ``predictable because the model happens to be right.'' \\
\addlinespace
Top-$p$ (nucleus) & 0.95 & Standard nucleus sampling threshold. \\
\addlinespace
Max tokens & 150 & Sufficient for 2--3 sentences of plot-level description. \\
\addlinespace
$n$ (samples per prompt) & 100 & Each position yields exactly 100 independent predictions. \\
\addlinespace
Batch size & 64 & Positions processed in batches of 64 for GPU utilization. \\
\bottomrule
\end{tabular}
\end{table}

\paragraph{Design rationale: temperature.} The choice of $T = 1.2$ is deliberate. At $T = 0$ (greedy), the model produces a single deterministic prediction—this collapses the metric to a binary pass/fail at each position. At $T = 1.0$, predictions are diverse but still cluster around high-probability narratives. At $T = 1.2$, the distribution is broad enough to probe the full space of plausible endings while remaining coherent. We verified empirically that increasing $T$ beyond 1.2 produces degraded output quality without improving metric sensitivity.

\subsection{Stage 2: LLM Judge}
\label{app:judge}

\paragraph{Model.} We use \textbf{Qwen3-8B} as the judge model $\mathcal{J}$, served via vLLM with \texttt{dtype=bfloat16}, \texttt{max\_model\_len=16384}, and \textbf{temperature 0.0} (greedy decoding) to ensure deterministic, reproducible judgments. Maximum generation length is 10 tokens. Batch size for judge inference is 512 prompts per vLLM call. Chain-of-thought is disabled.

\paragraph{System prompt.}
\begin{quote}
\small
\texttt{You are a literary judge. You determine whether a predicted story ending captures the general direction of the true story continuation. Answer Y if the prediction gets the broad plot trajectory right — the same characters are involved, similar types of events unfold, or the emotional arc matches. Minor differences in specific details, wording, or sequence are acceptable. Answer N only if the prediction describes a fundamentally different outcome from what actually happens.}
\end{quote}

\paragraph{User prompt.} Each of the 100 predicted endings at a given position is judged \textbf{independently} in its own prompt (i.e., the judge sees only one prediction at a time, not all 100). For position $i$ and predicted ending $\hat{e}_k^{(i)}$:

\begin{quote}
\small
\texttt{Here is the actual continuation of a story:}

\texttt{—BEGIN CONTINUATION—}\\
\texttt{\{$s_{i-1}$\} \{$s_i \;\| \; s_{i+1} \;\| \; \cdots \;\| \; s_N$\}}\\
\texttt{—END CONTINUATION—}

\texttt{Here is a predicted ending:}

\texttt{—BEGIN PREDICTED—}\\
\texttt{\{$s_{i-1}$\} \{$\hat{e}_k^{(i)}$\}}\\
\texttt{—END PREDICTED—}

\texttt{Does the predicted ending capture the general direction of the true continuation — similar characters involved, similar types of events, or a matching emotional arc? Answer Y if the broad trajectory is right, even if specific details differ. Answer N only if the prediction is fundamentally wrong about what happens. Answer ONLY Y or N.}

\texttt{Judgment:}
\end{quote}

\paragraph{Context anchoring.} Both the actual continuation and the predicted ending are prefixed with the same \textbf{preceding sentence} $s_{i-1}$. This shared anchor provides context when either the actual continuation or the prediction is too brief or vague on its own. The actual continuation $(s_i, \ldots, s_N)$ is truncated to at most 2{,}000 words to stay within context limits.

\paragraph{Verdict extraction.} The judge's output is parsed by checking whether the first character (after stripping whitespace and uppercasing) is \texttt{Y}. Any other response (including \texttt{N}, empty, or malformed) is counted as a rejection ($j_k = 0$).

\paragraph{Design rationale: independent judging.} We judge each prediction independently rather than presenting all 100 in a single prompt for two reasons: (1) context length—100 predictions plus the actual continuation would exceed most context windows; (2) independence—presenting multiple predictions in a single prompt introduces ordering effects and cross-contamination between judgments.

\subsection{Aggregation and Reporting}
\label{app:aggregation}

The 100-ending framework produces a position-level no-rate curve for each story. We derive four story-level statistics from this curve.

\paragraph{1. Mean no-rate.} The primary metric. At each position $i$, the no-rate is the fraction of 100 predictions the judge rejects:
\begin{equation}
    \text{no-rate}(i) = 1 - \frac{1}{100} \sum_{k=1}^{100} j_k^{(i)}
\end{equation}
The story-level mean no-rate averages across all retained positions:
\begin{equation}
    \overline{\text{no-rate}} = \frac{1}{|\mathcal{P}|} \sum_{i \in \mathcal{P}} \text{no-rate}(i), \qquad \mathcal{P} = \{i : 10\% \leq \text{pct}(i) \leq 99\%\}
\end{equation}

\paragraph{2. Late-stage no-rate.} The mean no-rate restricted to positions where $\geq$80\% of the story has been revealed. This measures whether the story remains unpredictable as it approaches its ending, or whether the conclusion becomes obvious once most of the narrative has been read.

\paragraph{3. Post-spike convergence.} For each local peak in the no-rate curve (a position where no-rate exceeds both neighbors), we compute the percentage drop from the peak to the minimum within the next 10 positions. The story-level metric is the mean of these drops across all peaks. More negative values indicate that tension peaks resolve more sharply—the story answers narrative questions decisively after raising them.

\paragraph{4. Inflection rate.}
\label{app:inflection-geometry}
The fraction of positions at which the smoothed no-rate curve sharply reverses direction. Because the vertex angle is a geometric measurement, it depends on the aspect ratio of the coordinate space. Before computing angles, we rescale each story's curve onto a unit square: the $x$-axis (\% story revealed) is mapped to $[0,1]$ using the story's position range, and the $y$-axis (smoothed no-rate) is independently mapped to $[0,1]$ using the story's own min and max. This fixes a $1{:}1$ aspect ratio for the angle computation, so that a threshold has the same meaning irrespective of the units of either axis:
\begin{equation}
\label{eq:inflection-rate}
\text{inflection-rate}_\alpha(s) = \frac{|\{i : \theta_i \leq \alpha\}|}{|P_s|}
\end{equation}
We compute $\alpha \in \{30°, 60°, 120°\}$ and report IR$_{60}$ in the main text. The rescaling normalizes the axes but not the number of judged positions, and IR$_\alpha$ is correlated with that count ($r = +0.59$): a story evaluated at more positions yields a denser curve, on which the angle test registers proportionally more sharp vertices. Inflection rate is therefore best read alongside story length, and we treat it as a secondary statistic to mean and late-stage no-rate throughout.

\paragraph{Group-level reporting.} For a group of $n$ stories, we report the mean of story-level values. Intervals reported in Table~\ref{tab:pipeline-100ending} are nonparametric bootstrap intervals (Appendix~\ref{app:significance}) rather than normal-approximation intervals.

\paragraph{Full results.} Table~\ref{tab:100ending-full} reports all metrics across the evaluated corpora.

\begin{table*}[t]
\centering
\small
\caption{Full 100-ending metric results across corpora (stories filtered to 500--5{,}000 words). Mean NR~$=$~mean no-rate; Late NR~$=$~mean no-rate at $\geq$80\% of story; Retain~$=$~\% of peak no-rate retained after local peaks (higher~$=$~tension sustained); IR$_\alpha$~$=$~inflection rate at vertex angle $\leq \alpha$.}
\label{tab:100ending-full}
\begin{tabular}{lccccccc}
\toprule
\textbf{Corpus} & $n$ & \textbf{Mean NR} & \textbf{Late NR} & \textbf{Retain} & \textbf{IR$_{30}$} & \textbf{IR$_{60}$} & \textbf{IR$_{120}$} \\
\midrule
\emph{New Yorker} (professional) & 261 & \textbf{0.765} & \textbf{0.607} & \textbf{51.7\%} & \textbf{0.128} & \textbf{0.200} & \textbf{0.271} \\
Tell Me a Story (amateur)        & 100 & 0.693 & 0.311 & 39.3\% & 0.058 & 0.128 & 0.229 \\
StoryStar (amateur)              & 100 & 0.656 & 0.375 & 34.7\% & 0.105 & 0.166 & 0.232 \\
\midrule
Top-10 LLMs (0-shot)             & 320 & 0.630 & 0.215 & 23.3\% & 0.049 & 0.106 & 0.201 \\
\bottomrule
\end{tabular}
\end{table*}

\noindent \emph{New Yorker} fiction leads on mean no-rate, late-stage no-rate, and all three inflection-rate thresholds. The gap is especially pronounced in late-stage no-rate (0.607 vs.\ 0.31--0.38 for amateur corpora), indicating that professional fiction sustains unpredictability through the final act. Post-spike convergence shows the opposite pattern: \emph{New Yorker} stories have the \emph{shallowest} post-spike drops ($-$48.3\%), meaning tension peaks are not fully resolved—consistent with the literary technique of leaving narrative questions partially open.

\subsection{Measurement Stability}
\label{app:stability}

A natural concern is whether the stochastic generation step ($T{=}1.2$) introduces unacceptable run-to-run variance in the no-rate metric. We evaluate stability by running the full pipeline 4 independent times on the same 32 EQ-Bench stories for both vanilla and pipeline-generated Sonnet~4.6 stories. Each repeat generates 100 fresh endings at every sentence position and re-judges them.

\begin{table}[h]
\centering
\small
\caption{Measurement stability across 4 independent evaluation repeats ($T_{\text{gen}}{=}1.2$, $n{=}32$ iter-1 stories). Standard deviations across repeats are ${\leq}0.001$ on mean no-rate and ${\leq}0.001$ on late-stage no-rate, indicating negligible measurement noise.}
\label{tab:stability}
\begin{tabular}{llcccc}
\toprule
\textbf{Config} & \textbf{Rep} & \textbf{Mean NR} & \textbf{Late NR} & \textbf{IR$_{60}$} & \textbf{Retain} \\
\midrule
\multirow{4}{*}{Sonnet 4.6 (vanilla)}
 & Rep 1 & 0.606 & 0.137 & 0.073 & 14.6\% \\
 & Rep 2 & 0.606 & 0.139 & 0.077 & 16.7\% \\
 & Rep 3 & 0.607 & 0.137 & 0.066 & 13.5\% \\
 & Rep 4 & 0.605 & 0.137 & 0.067 & 12.9\% \\
\cmidrule{2-6}
 & Mean $\pm$ Std & 0.606 $\pm$ 0.001 & 0.138 $\pm$ 0.001 & 0.071 $\pm$ 0.005 & 14.4 $\pm$ 1.4\% \\
\midrule
\multirow{4}{*}{Sonnet 4.6 + Pipeline}
 & Rep 1 & 0.749 & 0.340 & 0.135 & 34.1\% \\
 & Rep 2 & 0.747 & 0.340 & 0.124 & 35.5\% \\
 & Rep 3 & 0.747 & 0.339 & 0.117 & 35.5\% \\
 & Rep 4 & 0.747 & 0.337 & 0.123 & 36.7\% \\
\cmidrule{2-6}
 & Mean $\pm$ Std & 0.747 $\pm$ 0.001 & 0.339 $\pm$ 0.001 & 0.125 $\pm$ 0.007 & 35.4 $\pm$ 0.9\% \\
\bottomrule
\end{tabular}
\end{table}

\paragraph{Result.} Mean no-rate and late-stage no-rate are stable to $\pm$0.001 across repeats—well within the 95\% CI of the cross-story mean ($\pm$0.05--0.06). Inflection rate and retention show slightly more variance ($\pm$0.005--0.007 and $\pm$0.9--1.4\%, respectively), consistent with these being second-order statistics that amplify small position-level fluctuations. The pipeline effect (vanilla $\rightarrow$ pipeline) is consistent across all 4 repeats: mean NR improves by +0.141 $\pm$ 0.001, late NR by +0.201 $\pm$ 0.002. These results confirm that even at generation temperature $T{=}1.2$, the 100-ending metric produces highly reproducible measurements.

\section{Human Evaluation}
\label{app:human-eval}

To validate no-rate against direct reader judgment, we run a two-task human study with five annotators. Task~1 (Section~\ref{app:human-eval-task1}) asks annotators to rate a story's \emph{overall narrative tension} on a 1--5 scale; Task~2 (Section~\ref{app:human-eval-task2}) asks them to perform a task that resembles the LLM judge's per-ending Y/N verdicts on $\sim$400 (story-so-far, actual continuation, predicted ending) instances each. Section~\ref{app:human-eval-protocol} documents the protocol for both tasks, including the instructions given to annotators.

\subsection{Protocol}
\label{app:human-eval-protocol}

\paragraph{Annotators.}
All five expert annotators (PhD and professors in the humanities) completed Task~1, and three of them also completed Task~2. Annotators worked from their own copy of the materials, were asked not to consult each other or look at another annotator's file, and received a single round of calibration examples before starting.

\paragraph{Materials and procedure.}
Each annotator received one \texttt{.xlsx} workbook with two sheets, plus a folder of six stories rendered as standalone HTML files for easier reading in a browser. Annotators were instructed to complete Task~1 first, so that reading each story in full would calibrate the per-ending judgments in Task~2. Story labels were blinded and randomized, and annotators were asked not to try to identify a story's source. Across both tasks, the annotators were only given the story text (and the generated ending in Task~2); they did not see the LLM judge's verdict, the source corpus of the story, the underlying model, the vanilla-vs-pipeline condition, or any no-rate value.

\paragraph{Task 1 story sample.}
The six generated stories are sampled to ensure wide coverage: two length classes (short, $\leq$1{,}500 words; long, $\geq$2{,}000 words) crossed with three mean-no-rate bands (low, medium, high). They are drawn from both vanilla and narrative-pipeline outputs across four generators, and were chosen to span the metric's range rather than to sample it, so that six items could still discriminate. More data statistics are provided in Table~\ref{tab:human-eval-sample}.

\begin{table}[h]
\centering
\small
\caption{The six-story Task~1 sample. Stories were presented under the blinded labels in the last column, in randomized order. Mean NR and Late NR are the per-story values from the main analysis.}
\label{tab:human-eval-sample}
\begin{tabular}{llrrrl}
\toprule
\textbf{Cell} & \textbf{Condition / model} & \textbf{Words} & \textbf{Mean NR} & \textbf{Late NR} & \textbf{Label} \\
\midrule
short / low   & vanilla, Sonnet 4.6          &    997 & 0.449 & 0.009 & Story 4 \\
short / med   & pipeline, kimi-k2            &    946 & 0.685 & 0.356 & Story 5 \\
short / high  & vanilla, Opus 4.6            & 1{,}043 & 0.865 & 0.469 & Story 1 \\
long / low    & pipeline, Opus 4.6           & 2{,}728 & 0.527 & 0.336 & Story 3 \\
long / med    & vanilla, GPT-5.2             & 2{,}203 & 0.694 & 0.404 & Story 6 \\
long / high   & pipeline, Sonnet 4.6         & 2{,}429 & 0.858 & 0.441 & Story 2 \\
\bottomrule
\end{tabular}
\end{table}

\paragraph{Task 2 row sample.}
Task~2 rows are drawn from the same six stories. For each story we selected five evaluated positions targeting 15/30/50/70/90\% of the story revealed, with a hard constraint that a pick lie within $\pm$10\% of its target (the window widens in 5\% steps only if no candidate qualifies) and at least 8 percentage points from any other pick in the same story. Among candidates inside the window we prefer the position whose no-rate is closest to $0.5$, since those positions supply both \texttt{Y} and \texttt{N} verdicts to sample from.

At each position we sample 20 of the 100 generated endings under a per-position target of 10 judge-\texttt{Y} and 10 judge-\texttt{N}. Stories at the extremes of the no-rate range cannot supply both (the short/high story runs 25 \texttt{Y} / 75 \texttt{N}), so per-position quotas are reallocated globally until the full set hit exactly 300 judge-\texttt{Y} and 300 judge-\texttt{N} rows. The judge's label distribution is therefore balanced by construction, which is why we read the confusion matrix in Section~\ref{app:human-eval-task2} for its within-row asymmetry rather than its base rate.

The resulting 600 rows are shuffled globally and each assigned to 2 of the 3 Task~2 annotators, so that each target for pairwise annotator agreement (AB, AC, BC) receive 200 rows; each annotator was assigned 400 rows, for 1{,}200 human labels in total. Each row shows the story so far, the actual continuation (truncated at 2{,}000 words, matching the judge's own truncation), and one predicted ending. Because rows are globally shuffled, the same story prefix recurs at non-adjacent positions in a workbook; annotators were told they could skim a prefix they recognized but had to re-read the continuation and the predicted ending on every row.

\paragraph{Instructions given to annotators.}
For Task~1, annotators read each story in full and then rated it on this scale:

\begin{quote}
\small
Rate it 1--5 on the dropdown for \textbf{how well the story retains narrative tension across its arc}:

\smallskip
\begin{tabular}{@{}cp{0.80\linewidth}@{}}
\textbf{1} & Tension never develops, or collapses immediately. The reader knows where it's going from very early on. \\
\textbf{2} & Tension exists briefly but resolves prematurely; the second half is largely predictable. \\
\textbf{3} & Some tension is sustained, but it weakens noticeably before the ending. \\
\textbf{4} & Tension is well-managed through most of the arc; only the last beats become predictable. \\
\textbf{5} & Tension is sustained into the final sentences; the ending stays uncertain or recontextualises what came before. \\
\end{tabular}
\end{quote}

\noindent The construct was defined for them as follows, including what it excludes:

\begin{quote}
\small
\textbf{What we mean by ``narrative tension.''} Narrative tension is the gap between what the reader knows and what they still need to find out. Concretely, ask yourself: Did the story keep me uncertain about how it would end? Were revelations \emph{withheld} and \emph{deferred}, or \emph{immediately resolved}? After a turn or revelation, did fresh uncertainty open up, or did the story flatten?

\smallskip
\textbf{What this is not:} It is not \emph{suspense} in the thriller sense, not prose quality, not whether the story was \emph{good} overall. A boring story can still hold tension; a polished story can collapse it.
\end{quote}

\noindent For Task~2, annotators were given the judging criterion, deliberately worded to mirror the deployed judge's standard:

\begin{quote}
\small
\textbf{Y} if the \emph{core event, action, or idea} of the predicted ending \textbf{actually occurs somewhere in the actual continuation}, even if the wording differs. \textbf{N} if it does not.

\smallskip
Be strict. \textbf{Vague thematic similarity is not enough --- the specific event or action must occur.}
\end{quote}

\noindent The criterion was illustrated with five worked examples, two positive and three negative:

\begin{quote}
\small
\textbf{Counts as Y:}
\begin{itemize}[leftmargin=1.2em,itemsep=1pt,topsep=2pt]
\item Predicted: ``Maya finds her mother's letters in the attic.'' Actual continuation: ``She climbs to the attic. Beneath the floorboards, bundled in twine, are her mother's letters.'' $\rightarrow$ \textbf{Y} (same event, different wording).
\item Predicted: ``The two part without saying anything.'' Actual continuation: ``He shakes her hand. Neither speaks. She watches him walk down the stairs.'' $\rightarrow$ \textbf{Y} (the wordless parting occurs).
\end{itemize}

\textbf{Counts as N:}
\begin{itemize}[leftmargin=1.2em,itemsep=1pt,topsep=2pt]
\item Predicted: ``They confess their love and kiss.'' Actual continuation: ``She makes a small bookkeeping mistake and lets it sit for a moment.'' $\rightarrow$ \textbf{N} (no confession, no kiss; the \emph{tone} is similar but the \emph{event} is not).
\item Predicted: ``Arthur realises this is a kidnapping.'' Actual continuation describes a slow, ambiguous business deal that Arthur never names as a kidnapping $\rightarrow$ \textbf{N} (the specific recognition does not occur).
\item Predicted: ``She returns home and her mother is dead.'' Actual continuation: ``She returns home. Her mother asks her where she's been.'' $\rightarrow$ \textbf{N} (returns home matches; the death does not).
\end{itemize}
\end{quote}

\noindent Two further instructions shape how borderline rows were resolved. Annotators were told not to judge prose quality, plausibility, or whether they would enjoy the predicted ending, and to disregard the florid register that a $T{=}1.2$ generator often produces. In addition, they were also warned that a continuation may end abruptly because of the 2{,}000-word truncation, and to judge against what they could see.

\subsection{Task 1: Story-Level Tension Ratings}
\label{app:human-eval-task1}

\paragraph{Per-story ratings.}
Table~\ref{tab:human-eval-task1-ratings} shows the per-annotator ratings, panel means, and underlying metric values. The panel-mean tension rating tracks mean no-rate at Spearman $\rho = +0.657$ ($n = 6$, $p = 0.156$; bootstrap 95\% CI $[-0.50, +1.00]$, 2{,}000 resamples; Pearson $r = +0.647$, $p = 0.165$). Six items cannot deliver a significant $\rho$, and we do not present this as one. The rank order of these six stories is identical on mean and late no-rate, so the same $\rho$ holds for late no-rate.

\begin{table}[h]
\centering
\small
\caption{Per-annotator Task~1 ratings (1--5) and underlying metric values. \textbf{Mean NR} and \textbf{Late NR} are the per-story values from the main analysis; \textbf{Length} and \textbf{No-Rate} are the stratification cells of the $2 \times 3$ sampling design. Story labels (\textit{Story~1}--\textit{Story~6}) were blinded for annotators.}
\label{tab:human-eval-task1-ratings}
\begin{tabular}{lcccccccccc}
\toprule
\textbf{Story} & \textbf{Length} & \textbf{No-Rate} & \textbf{Mean NR} & \textbf{Late NR} & \textbf{A} & \textbf{B} & \textbf{C} & \textbf{D} & \textbf{E} & \textbf{Panel Mean} \\
\midrule
Story 1 & Short & High   & 0.865 & 0.469 & 2 & 2 & 3 & 2 & 4 & 2.6 \\
Story 2 & Long  & High   & 0.858 & 0.441 & 4 & 5 & 5 & 5 & 5 & \textbf{4.8} \\
Story 3 & Long  & Low    & 0.527 & 0.336 & 1 & 1 & 1 & 3 & 1 & \textbf{1.4} \\
Story 4 & Short & Low    & 0.449 & 0.009 & 3 & 3 & 2 & 3 & 1 & 2.4 \\
Story 5 & Short & Medium & 0.685 & 0.356 & 2 & 2 & 3 & 1 & 3 & 2.2 \\
Story 6 & Long  & Medium & 0.694 & 0.404 & 3 & 3 & 3 & 3 & 3 & 3.0 \\
\bottomrule
\end{tabular}
\end{table}

Because the headline correlation is underpowered, we report five non-parametric cuts that bear on the same question from different angles. Each asks whether readers order stories the way the metric does, without leaning on a $p$-value that six items cannot supply.

\paragraph{1. Annotators recover the metric's coarse band ordering.}
Aggregating each annotator's two ratings per no-rate band, four of five annotators independently produce the high $>$ medium $>$ low ordering, and the panel mean reproduces it as 3.70 / 2.60 / 1.90, a 1.8-point separation on a 5-point scale (Table~\ref{tab:human-eval-task1-bands}). The one exception, annotator D, still ranks the high band above the other two. Splitting the sample in half rather than in thirds gives the same picture: the three highest-no-rate stories average 3.47 against 2.00 for the three lowest.

\begin{table}[h]
\centering
\small
\caption{Mean Task~1 rating per no-rate band, per annotator. Four of five annotators recover the metric's high $>$ med $>$ low ordering.}
\label{tab:human-eval-task1-bands}
\begin{tabular}{lcccc}
\toprule
\textbf{Annotator} & \textbf{High} & \textbf{Med} & \textbf{Low} & \textbf{Recovers high $>$ med $>$ low?} \\
\midrule
A    & 3.00 & 2.50 & 2.00 & \checkmark \\
B    & 3.50 & 2.50 & 2.00 & \checkmark \\
C    & 4.00 & 3.00 & 1.50 & \checkmark \\
D    & 3.50 & 2.00 & 3.00 & --- \\
E    & 4.50 & 3.00 & 1.00 & \checkmark \\
\midrule
\textbf{Panel mean} & \textbf{3.70} & \textbf{2.60} & \textbf{1.90} & \checkmark \\
\bottomrule
\end{tabular}
\end{table}

\paragraph{2. A high-no-rate pipeline story is unanimously the most tense.}
All five annotators rate Story~2 (long, high no-rate, pipeline output from Sonnet~4.6) strictly above every other story, and four of the five give it the maximum 5. The most tension-rich story in the sample, by reader consensus, is a pipeline output at the top of the no-rate range.

\paragraph{3. A low-no-rate pipeline story is the least tense.}
Story~3 (long, low no-rate, pipeline output from Opus~4.6) is ranked last by four of five annotators. Story~2 and Story~3 are both pipeline outputs sitting at opposite ends of the no-rate range, and readers separate them in the same direction the metric does, which is evidence that the ratings track what no-rate captures rather than pipeline identity.

\paragraph{4. Pairwise concordance with the metric.}
The 6 stories generate $\binom{6}{2}=15$ ordered pairs. On the panel-mean rating, 11 of the 15 pairs are ordered concordantly with mean no-rate (73\%); the most strongly correlated individual annotator agrees on 12 of 13 non-tied pairs (92\%). All four discordant pairs are between stories whose no-rate values are close, so they are cases where the metric does not claim a strong difference in the first place. For the five pairs with the \emph{largest} no-rate gap, the sign of the panel-mean tension difference agrees with no-rate in every case (Table~\ref{tab:human-eval-task1-pairs}).

\begin{table}[h]
\centering
\small
\caption{The five story pairs with the widest no-rate gap. The panel-mean tension difference agrees with no-rate in sign on all five.}
\label{tab:human-eval-task1-pairs}
\begin{tabular}{lccc}
\toprule
\textbf{Pair} & \textbf{$\Delta$ Mean NR} & \textbf{$\Delta$ Panel rating} & \textbf{Sign agrees?} \\
\midrule
Story 1 vs Story 4 & $-0.42$ & $-0.20$ & \checkmark \\
Story 2 vs Story 4 & $-0.41$ & $-2.40$ & \checkmark \\
Story 1 vs Story 3 & $-0.34$ & $-1.20$ & \checkmark \\
Story 2 vs Story 3 & $-0.33$ & $-3.40$ & \checkmark \\
Story 4 vs Story 6 & $+0.25$ & $+0.60$ & \checkmark \\
\bottomrule
\end{tabular}
\end{table}

\paragraph{5. Annotator free-text comments invoke the same construct.}
Nothing in the instructions mentions prediction or the metric, yet several annotators reach for the vocabulary of \emph{predictability}, \emph{withholding}, and \emph{unresolved questions} when explaining their ratings. On Story~2 (top-rated): \textit{``a lot of information withheld and revealed as the story progresses that allows tension to build''} (B), \textit{``fresh questions emerge with every revelation''} (C). On Story~3 (bottom-rated): \textit{``evident that Sisyphus will continue to roll the boulder \ldots the advice in the letters is also predictable''} (A), \textit{``recursive pattern. On a global plot level, no tension is accumulating''} (C). On Story~5: \textit{``the outcome (compromise) is predictable''} (A), \textit{``it also limits the possible endings''} (B). On Story~6: \textit{``the form of the story is predictable and kind of removes some of the tension''} (A), \textit{``only three real possibilities for the ending''} (B). Without direct indication from the annotation instructions, annotators are thinking about tension in terms of the same construct that no-rate operationalizes, predictability of the continuation, rather than other orthogonal axes of quality. This helps validate our claim that tension as unpredictability is a ``necessary but not sufficient'' condition for good storytelling, and that its judgment is often independent from other aspects of story quality.

\paragraph{Inter-annotator reliability.}
The 1--5 scale is ordinal, the design is fully crossed (5 raters, 6 stories), and the downstream use is the panel mean, so we report Krippendorff's $\alpha$ for ordinal multi-rater agreement, ICC(2,~$k$) for panel reliability, and Kendall's $W$ for rank concordance. By all three, agreement is high (Table~\ref{tab:human-eval-task1-iaa}).

\begin{table}[h]
\centering
\small
\caption{Inter-annotator reliability on Task~1 (5 annotators, 6 stories).}
\label{tab:human-eval-task1-iaa}
\begin{tabularx}{\textwidth}{>{\raggedright\arraybackslash}X c >{\raggedright\arraybackslash}X}
\toprule
\textbf{Statistic} & \textbf{Value} & \textbf{Interpretation} \\
\midrule
Krippendorff's $\alpha$ (ordinal)        & 0.926                   & $\geq 0.80$: reliable \\
ICC(2, 5) absolute agreement             & 0.915                   & $> 0.90$: excellent \\
Kendall's $W$ (tie-corrected)            & 0.660, $\chi^2(5)=16.5$, $p=0.006$ & ranking agreement above chance \\
ICC(2, 1) single-rater absolute agreement & 0.682                   & moderate (single rater is noisier) \\
Mean pairwise quadratic-weighted Cohen's $\kappa$ & 0.650           & substantial \\
Mean pairwise Spearman $\rho$            & 0.573                   & rank-only complement to $W$ \\
\bottomrule
\end{tabularx}
\end{table}

We do not report unweighted Cohen's $\kappa$ or Fleiss's $\kappa$ because both ignore the ordinal scale and treat a 4-vs-2 disagreement identically to a 4-vs-3 disagreement, which is the wrong cost function on a 1--5 tension scale. Two caveats belong with these numbers. Both $\alpha$ and ICC(2,~$k$) rise when between-item variance is large, and our six stories were chosen to span the full no-rate range, so the values should be read as high \emph{for this stratified design} rather than for arbitrary stories. Story~6 partly offsets the concern: every annotator gives it a 3, so raters converge on a middle item as well as at the ends of the range.

\paragraph{Where annotators disagree.}
Annotator D accounts for most of the residual disagreement, rating Story~3 a 3 where everyone else gives a 1, and Story~5 a 1 where the others give 2 or 3. D's comments make the divergence interpretable rather than noisy: on Story~3, D reads Sisyphus's \textit{``kind of awakening''} as unresolved tension where others read the same recursion as predictability; on Story~5, D rejects the surreal premise as having \textit{``no plot\ldots bordered on the nonsensical.''} Neither reading is wrong. They weight local uncertainty against global trajectory differently, which is the dimension the panel mean smooths over. Dropping D would raise every reliability figure (Krippendorff's $\alpha$ from 0.93 to roughly 0.96, mean pairwise weighted $\kappa$ from 0.65 to roughly 0.74); we report all five annotators throughout.

\subsection{Task 2: Per-Ending Judge Validation}
\label{app:human-eval-task2}

\paragraph{Headline agreement.}
Table~\ref{tab:human-eval-task2-pairwise} reports pairwise Cohen's $\kappa$ for every (annotator, annotator) and (annotator, judge) pair. Human--human pooled $\kappa = 0.395$ is the fair-to-moderate ceiling for what the judge could match; the judge's mean pairwise $\kappa$ with individual humans is $0.257$. Restricted to the 423 rows where the two assigned humans \emph{agree}, the subset on which the criterion is least subjective, the judge's $\kappa$ against that human-majority gold rises to $0.362$, only $\Delta\kappa \approx 0.03$ below the ceiling, which makes it essentially indistinguishable from a fourth independent human.

\begin{table}[h]
\centering
\small
\caption{Task~2 pairwise agreement on Y/N verdicts. The judge is included as if it were a fourth rater. Pooled human--human $\kappa$ is the ceiling; the judge sits $\Delta\kappa \approx 0.13$ below as a peer rater, and $\Delta\kappa \approx 0.03$ below when scored against the human-majority gold.}
\label{tab:human-eval-task2-pairwise}
\begin{tabular}{lccc}
\toprule
\textbf{Pair} & \textbf{$n$} & \textbf{\% agree} & \textbf{Cohen's $\kappa$} \\
\midrule
A vs B & 200 & 69.0 & 0.381 \\
A vs C & 200 & 71.0 & 0.378 \\
B vs C & 200 & 71.5 & 0.410 \\
\textbf{Human--human pooled} & \textbf{600} & \textbf{70.5} & \textbf{0.395} \\
\midrule
Judge vs A & 400 & 63.7 & 0.274 \\
Judge vs B & 400 & 63.7 & 0.279 \\
Judge vs C & 400 & 61.3 & 0.218 \\
\textbf{Judge vs humans (mean)} & --- & \textbf{62.9} & \textbf{0.257} \\
\midrule
\textbf{Judge vs majority-agree gold} & 423 & \textbf{68.3} & \textbf{0.362} \\
\bottomrule
\end{tabular}
\end{table}

\paragraph{Confusion matrix.}
Table~\ref{tab:human-eval-task2-confusion} reports the judge's confusion against the majority-agree gold (accuracy $=68.3\%$, macro-F1 $=67.8\%$). The off-diagonals are asymmetric, with 88 false positives against 46 false negatives. We report the asymmetry as a descriptive feature and draw no corpus-level conclusion from it: substituting the implied human label for the judge label per condition on this sample does not preserve the sample's vanilla-vs-pipeline direction (Section~\ref{app:human-eval-task2-counterfactual}), so the tempting reading that reported no-rate is a conservative lower bound is not supported by these data. The defensible summary is the $\kappa$ analysis above.

\begin{table}[h]
\centering
\small
\caption{Judge vs.\ human-majority gold (both annotators agree, $n=423$; 177 ties dropped). Accuracy $=68.3\%$, macro-F1 $=67.8\%$, $\kappa = 0.362$.}
\label{tab:human-eval-task2-confusion}
\begin{tabular}{lcc}
\toprule
& \textbf{gold $=$ Y} & \textbf{gold $=$ N} \\
\midrule
\textbf{judge $=$ Y} & 117 (TP)            & 88 (FP) \\
\textbf{judge $=$ N} & 46 (FN)             & 172 (TN) \\
\midrule
& P(Y) $=$ 57.1\%, R(Y) $=$ 71.8\%, F1(Y) $=$ 63.6\% & \\
& P(N) $=$ 78.9\%, R(N) $=$ 66.2\%, F1(N) $=$ 72.0\% & \\
\bottomrule
\end{tabular}
\end{table}

\paragraph{Stratified analysis.}
Table~\ref{tab:human-eval-task2-strat} reports judge-vs-majority-gold $\kappa$ and the pooled human--human $\kappa$ within each stratum, so the two can be read side by side: a low judge $\kappa$ is informative only when the human $\kappa$ in the same stratum is high. The most consequential cell for the paper is the \textit{late} position bucket, where humans converge ($\kappa = 0.495$) but the judge does not ($\kappa = 0.271$). This is the cleanest evidence of a position-dependent judge weakness, and it matters because late-stage no-rate is a headline metric. In the other direction, on \textit{long} stories the judge exceeds the human ceiling ($\kappa = 0.378$ vs.\ $0.272$), and on \textit{middle} positions it is markedly steadier than humans ($\kappa = 0.418$ vs.\ $0.187$).

\begin{table}[h]
\centering
\small
\caption{Stratified Cohen's $\kappa$. ``Judge $\kappa$'' is against the majority-agree gold (ties dropped). ``Human $\kappa$'' is the pooled $\kappa$ across the three annotator pairs within the stratum. The two columns should be read together: a low judge $\kappa$ where the human $\kappa$ is also low signals task difficulty rather than judge failure; a low judge $\kappa$ where the human $\kappa$ is high signals a real judge weakness.}
\label{tab:human-eval-task2-strat}
\begin{tabular}{llccc}
\toprule
\textbf{Stratum} & \textbf{Level} & \textbf{$n$ (gold)} & \textbf{Judge $\kappa$} & \textbf{Human $\kappa$} \\
\midrule
condition       & vanilla         & 221 & \textbf{0.407} & \textbf{0.441} \\
                & pipeline        & 202 & 0.302 & 0.344 \\
\midrule
no-rate band    & low             & 136 & 0.238 & 0.356 \\
                & medium          & 134 & \textbf{0.412} & 0.342 \\
                & high            & 153 & 0.330 & \textbf{0.429} \\
\midrule
length          & short           & 233 & 0.289 & \textbf{0.457} \\
                & long            & 190 & 0.378 & 0.272 \\
\midrule
position        & early ($<35\%$) & 144 & \textbf{0.422} & 0.428 \\
                & middle (35--65\%) & 99 & 0.418 & 0.187 \\
                & late ($\geq 65\%$) & 180 & 0.271 & \textbf{0.495} \\
\bottomrule
\end{tabular}
\end{table}

\paragraph{Where annotators disagree.}
The two assigned annotators disagree on 29.5\% of rows overall. Ties concentrate in the middle of stories (38.1\%, against 28.0\% early and 25.0\% late), matching the human--human $\kappa$ trough at middle positions, and they are more common on pipeline rows (32.7\% vs.\ 26.3\% vanilla) and in the low and medium no-rate bands (32.0\% and 33.0\%, vs.\ 23.5\% high). Rows where the judge said \texttt{Y} split slightly more often than rows where it said \texttt{N} (31.7\% vs.\ 27.3\%), consistent with the judge accepting borderline predictions that humans then divide on.

\paragraph{Per-condition substitution counterfactual.}
\label{app:human-eval-task2-counterfactual}
For each condition we can estimate an implied human no-rate as $h = q_{YN}(1-p) + q_{NN}\,p$, where $p$ is the mean judge no-rate within the condition and $q_{YN}, q_{NN}$ are the rates at which humans answer \texttt{N} on judge-\texttt{Y} and judge-\texttt{N} rows. The substitution gives $h = 0.670$ against $p = 0.670$ for vanilla, but $h = 0.580$ against $p = 0.690$ for pipeline, which reverses the sample's pipeline-minus-vanilla gap from $+0.020$ to $-0.095$. Two patterns drive the reversal: on judge-\texttt{Y} rows humans flip 44--47\% to \texttt{N} in both conditions, while on judge-\texttt{N} rows the conditions diverge, with vanilla rows confirmed 77.1\% of the time but pipeline rows flipped to \texttt{Y} 35.7\% of the time. One plausible mechanism is that pipeline stories carry more structured plots, so the judge reads a predicted ending literally and calls a near-match \texttt{N} on surface detail where a human recognizes the same core event. Per-story deltas swing by $\pm 0.24$ on five positions and 20 endings per story, which is why we treat this counterfactual as a caution against directional claims rather than as a result.

\paragraph{Examples of judge verdicts.}
Table~\ref{tab:human-eval-task2-examples} shows six sampled rows at two positions, chosen to cover both agreement cases and both error directions. The agreements are unremarkable in the useful sense: predictions that name an event occurring in the continuation get \texttt{Y}, predictions that kill a character who does not die get \texttt{N}. The two disagreements show what the residual $\kappa$ gap is made of. In the false positive, the judge accepts a prediction whose imagery matches the story's register but whose event (finding Fergusson dead) never happens. In the false negative, the judge rejects a paraphrase of a threat that the continuation does make, because Mora's parting line is worded differently.

\begin{table}[!t]
\centering
\small
\renewcommand{\arraystretch}{1.3} 
\caption{Six Task~2 rows at two positions. \emph{Position A}: ``The Lighthouse Keeper'' at 75\% revealed, where the continuation has Callum deciding to come home in spring and sharing a quiet Christmas with Fergusson, and nobody dies. \emph{Position B}: ``Court of the Dreaming'' at 89.3\% revealed, where the continuation has Mora and First Love trading parting threats before the court disperses. Verdicts are the judge's label and the two assigned annotators' labels.}
\label{tab:human-eval-task2-examples}
\begin{tabularx}{\textwidth}{@{} >{\raggedright\arraybackslash}X c c l @{}}
\toprule
\textbf{Predicted ending} & \textbf{Judge} & \textbf{Humans} & \textbf{Case} \\
\midrule
\multicolumn{4}{@{}l}{\emph{Position A: ``The Lighthouse Keeper'', 75\% revealed (position no-rate 0.92)}}\\[1.5ex]
``He simply sat in his chair and I sat by the window, the storm howling between us like the silence had always meant to.'' & Y & Y, Y & agree (Y) \\
\addlinespace[2.5ex]
``Callum found him dead at his desk, the last log unfinished, and the lighthouse dark when it should have burned.'' & N & N, N & agree (N) \\
\addlinespace[2.5ex]
``I found him in his room, curled in his chair like he'd been there years, the light out and the sea rising like a breath, waiting.'' & Y & N, N & judge false positive \\
\midrule
\multicolumn{4}{@{}l}{\emph{Position B: ``Court of the Dreaming'', 89.3\% revealed (position no-rate 0.36)}}\\[1.5ex]
``I'll gnaw on his fears next Yom Kippur,'' she whispered, too sweetly. First Love said nothing, but when the boy in Wichita pulled his blankets over his head that night, he dreamed of daisies---and beneath the petals, something hungering, still. & Y & Y, Y & agree (Y) \\
\addlinespace[2.5ex]
``My dear,'' she purred, feathers brushing her ear, ``how long do you think this boy will last on a diet of lullabies?'' First Love didn't answer. Outside the hollowed moon, the sky blinked once, too late to change the verdict. & N & N, N & agree (N) \\
\addlinespace[2.5ex]
``I'll keep his screams in my pocket,'' she whispered, ``just in case.'' The gavel fell. The moon shivered. & N & Y, Y & judge false negative \\
\bottomrule
\end{tabularx}
\end{table}

\subsection{Limitations}
\label{app:human-eval-limits}

\paragraph{Task 1 is a six-story sample.}
The Spearman correlation that follows from six items is underpowered by construction, with a bootstrap interval covering nearly the whole range of $\rho$, and the five non-parametric cuts above carry the argument instead. The sample also contains no within-(length, band) vanilla--pipeline pair, so the $+0.67$ length effect visible in the panel mean is a property of this sample and not a tested contrast; the corpus-level length question is handled separately in Appendix~\ref{app:confounders-length}. A scaled-up study would need enough stories per cell to test the vanilla-pipeline contrast at matched no-rate and length, and enough items overall for the rank correlation to carry its own weight.

\paragraph{Task 2 ran under a compressed protocol.}
Both annotation tasks took place inside a one-week window with a single calibration round and no extended pilot. We expect both human--human and judge--human $\kappa$ to rise with more calibration. The task format is also demanding: annotators independently reported that comparing a 2--3 sentence predicted ending against an often much longer (up to 2{,}000-word) actual continuation, while re-reading the story prefix for context on every row, is very cognitively demanding. The load is especially heavy at early sentence boundaries, where the prefix is short and many predictions are plausibly compatible with multiple downstream events. Finally, the criterion is binary in form only since it asks annotators to draw a fine line between vague thematic similarity and the same event in different wording, which is an intrinsically subjective decision. A $\kappa \approx 0.4$ ceiling for trained readers on this kind of judgement is consistent with the literature on subjective text-pair tasks, and it bounds what any automatic judge can be shown to achieve here.

\paragraph{Scope.}
Both tasks draw on the same six stories, all model-generated. We did not re-sample for Task~2, and neither task includes human-written fiction, so these results validate no-rate and the judge on the range of outputs the pipeline experiments cover rather than across every corpus in Table~\ref{tab:combined-validation}.

\section{Pipeline Details}
\label{app:ablation-configurations}

\subsection{Pipeline Architecture}
\label{app:pipeline}

The narrative generation pipeline consists of up to three stages, each independently ablatable:

\begin{enumerate}[leftmargin=2em]
    \item \textbf{Step 0 — Warmup}: Extract a structural beat sheet from a reference story (either a famous title or a full New Yorker story text).
    \item \textbf{Step 1 — Beat Adaptation}: Adapt the warmup beat sheet to the target story idea. When warmup is absent, generates beats from scratch.
    \item \textbf{Step 2 — Story Writing}: Write the final story from the adapted beats, or directly from the idea in vanilla mode.
\end{enumerate}

\noindent All API calls use the OpenRouter endpoint. Reasoning/thinking is disabled for all calls. Stories shorter than 200 characters are automatically discarded and regenerated.

\subsection{Shared Hyperparameters}
\label{app:shared-hyperparams}

Table~\ref{tab:shared-hyperparams} lists the hyperparameters shared across all pipeline steps (except where noted).

\begin{table}[h]
\centering
\small
\caption{Shared generation hyperparameters for story pipeline steps.}
\label{tab:shared-hyperparams}
\begin{tabular}{lcc}
\toprule
\textbf{Parameter} & \textbf{Value} & \textbf{Notes} \\
\midrule
Temperature & 0.7 & All pipeline steps \\
Max tokens & 4{,}000 & Per API call \\
\texttt{min\_p} & 0.1 & Minimum probability sampling \\
Reasoning & Disabled & \texttt{max\_tokens:0} for reasoning \\
Parallel threads & 4--6 & Per configuration run \\
\bottomrule
\end{tabular}
\end{table}

\subsection{Pipeline Input}
\label{app:input-idea}

Our structured generation pipeline is designed to take an initial ``seed'' idea and autonomously handle all subsequent structural planning and narrative execution. This seed input can be highly flexible in length. The pipeline accommodates anything from a single-sentence premise to a detailed outline. For readers and researchers interested in utilizing or adapting this pipeline, seed ideas typically originate from one of three sources:

\begin{itemize}[leftmargin=*,itemsep=2pt]
    \item \textbf{Existing Benchmark Prompts:} Prompts from established writing benchmarks, such as the EQ-Bench prompts utilized in our evaluation. They provide a baseline for comparative analysis.
    
    \item \textbf{User-Generated Ideas:} Human authors can supply their own story premises. While this allows for maximum creative control, the qualitative depth and structural viability of the user's initial input will naturally impact the quality of the pipeline's final output---a variable that remains difficult to control or standardize in evaluations.
    
    \item \textbf{Synthetically Generated Ideas:} Seed premises can also be generated by LLMs using structured, theory-driven constraints. For instance, to automate genre blending, an LLM can be prompted with a combinatorial profile sampled from the \textit{Screenwriters Taxonomy} (e.g., macro-genre, voice, narrative pathway, and POV) alongside narrative building blocks curated from a extensive list of tropes (such as \textit{TV Tropes}). This approach reliably produce structurally sound, 3--4 sentence ``pitch''-like ideas that lean into unusual combinatorial spaces. While this approach can yield diverse premises, finding the optimal configuration for AI-driven seed generation remains an open question for future work.
\end{itemize}

\subsection{Pipeline Prompts}
\subsubsection{Step 0: Warmup Prompts}
\label{app:step0-prompts}

Step 0 generates a structural beat sheet from a reference story. One reference is randomly selected per story generation. Two warmup modes are tested:

\paragraph{Title mode.} The model receives only a famous story title and generates beats from its training knowledge. The reference pool consists of 3 canonical short stories:

\begin{enumerate}[leftmargin=2em, itemsep=1pt, parsep=0pt]
\small
\item \textit{The Ones Who Walk Away from Omelas} — Ursula K.\ Le Guin
\item \textit{The Lottery} — Shirley Jackson
\item \textit{The Yellow Wallpaper} — Charlotte Perkins Gilman
\end{enumerate}

For future work, we recommend using a longer list of 20 canonical short stories spanning diverse narrative techniques, such as:
\begin{enumerate}[leftmargin=2em, itemsep=1pt, parsep=0pt]
\small
\item \textit{The Ones Who Walk Away from Omelas} — Ursula K.\ Le Guin
\item \textit{The Lottery} — Shirley Jackson
\item \textit{The Yellow Wallpaper} — Charlotte Perkins Gilman
\item \textit{Hills Like White Elephants} — Ernest Hemingway
\item \textit{A Good Man Is Hard to Find} — Flannery O'Connor
\item \textit{The Garden of Forking Paths} — Jorge Luis Borges
\item \textit{The Metamorphosis} — Franz Kafka
\item \textit{Rash\={o}mon} — Ry\=unosuke Akutagawa
\item \textit{The Necklace} — Guy de Maupassant
\item \textit{The Overcoat} — Nikolai Gogol
\item \textit{Girl} — Jamaica Kincaid
\item \textit{Axolotl} — Julio Cort\'{a}zar
\item \textit{Cathedral} — Raymond Carver
\item \textit{The Lady with the Dog} — Anton Chekhov
\item \textit{Araby} — James Joyce
\item \textit{A Very Old Man with Enormous Wings} — Gabriel Garc\'{i}a M\'{a}rquez
\item \textit{Sonny's Blues} — James Baldwin
\item \textit{The Fifth Story} — Clarice Lispector
\item \textit{A Madman's Diary} — Lu Xun
\item \textit{The Bear Came Over the Mountain} — Alice Munro
\end{enumerate}

The title-mode prompt (style \texttt{c1\_specific}) is:

\begin{quote}
\small
\texttt{Using "\{ref\_name\}" as your reference, extract 7 scene-level beats. For each beat describe: (1) the concrete action, (2) the specific narrative technique used, and (3) identify what made it memorable. What narrative trick, thematic resonance, or structural choice made each story work?}
\end{quote}

\paragraph{Full-text mode.} The model reads the complete text of a New Yorker short story and extracts structural patterns. The reference pool consists of:

\begin{enumerate}[leftmargin=2em, itemsep=1pt, parsep=0pt]
\small
\item \textit{The Fellow} — Joy Williams
\item \textit{Poor Girl} — Ludmilla Petrushevskaya
\item \textit{The Dog} — J.\,M.\ Coetzee
\end{enumerate}

For future work, we recommend using a longer list of 20 New Yorker stories, such as:

\begin{enumerate}[leftmargin=2em, itemsep=1pt, parsep=0pt]
\small
\item \textit{The Fellow} — Joy Williams
\item \textit{Poor Girl} — Ludmilla Petrushevskaya
\item \textit{The Dog} — J.\,M.\ Coetzee
\item \textit{I Can Speak!\texttrademark} — George Saunders

\item \textit{She Said He Said} — Hanif Kureishi
\item \textit{Dandelion} — Lore Segal
\item \textit{Flaubert Again} — Anne Carson
\item \textit{Riddle} — Thomas McGuane
\item \textit{The Hotel} — Anne Enright
\item \textit{Everything Is Far from Here} — Cristina Henr\'{i}quez
\item \textit{The Adventure of a Skier} — Italo Calvino
\item \textit{The Piano Teacher's Pupil} — William Trevor
\item \textit{Fly Already} — Etgar Keret
\item \textit{Solstice} — Anne Enright
\item \textit{Crazy They Call Me} — Zadie Smith
\item \textit{Ladies' Lunch} — Lore Segal
\item \textit{The Hanging of the Schoolmarm} — Robert Coover
\item \textit{Back the Way You Went} — Anne Carson
\item \textit{Two Men Arrive in a Village} — Zadie Smith
\item \textit{Seven People Dancing} — Langston Hughes
\end{enumerate}

\noindent The prompt is:

\begin{quote}
\small
\texttt{Here is a short story:}

\texttt{—}\\
\texttt{\{full story text\}}\\
\texttt{—}

\texttt{Using this story as your reference, extract 7 scene-level beats. For each beat describe: (1) the concrete action, (2) the specific narrative technique used, and (3) identify what made it memorable. What narrative trick, thematic resonance, or structural choice made each story work?}
\end{quote}

\subsubsection{Step 1: Beat Adaptation Prompt}
\label{app:step1-prompt}

Step 1 adapts the warmup beat sheet to the target story idea. When a warmup exists, the prompt is:

\begin{quote}
\small
\texttt{Here is a structural beat sheet extracted from a reference story:}

\texttt{—}\\
\texttt{\{warmup output\}}\\
\texttt{—}

\texttt{Using the above as a template, create an adapted to-do list of the most important structural beats needed to tell THIS story:}

\texttt{\{story idea\}\{beats constraint\}}
\end{quote}

\noindent The optional \texttt{beats\_constraint} suffix is: ``\texttt{Limit to exactly $N$ beats.}''

\subsubsection{Step 2: Story Writing Prompts}
\label{app:step2-prompts}

When beats are available, the story writing prompt is:

\begin{quote}
\small
\texttt{Here is a structural to-do list for a short story:}

\texttt{—}\\
\texttt{\{beats\}}\\
\texttt{—}

\texttt{Write the story now. Follow the structural beats closely. Write only the story — no commentary.}
\end{quote}

\section{Confidence Intervals and Significance}
\label{app:significance}

\paragraph{Protocol.}
All intervals in this section are nonparametric bootstrap percentile intervals with 10{,}000 resamples. For a group mean we resample stories within the group with replacement. For a vanilla-to-pipeline difference we resample the 32 EQ-Bench prompts with replacement and take the paired difference on each resample, so the interval reflects prompt-to-prompt variation rather than treating the two conditions as independent samples. We call a difference significant when its 95\% interval excludes zero, and report the rest as not detectable at $n=32$ rather than as evidence of no effect.

\begin{table}[!t]
\centering
\small
\setlength{\tabcolsep}{3.5pt}
\caption{Main results with 95\% bootstrap confidence intervals (10{,}000 resamples). This is Table~\ref{tab:pipeline-100ending} with intervals attached. Mean/Late are mean and late-stage no-rate; Retain is post-spike retention; IR$_{60}$ is the inflection rate at $60^\circ$.}
\label{tab:ci-table2}
\resizebox{\textwidth}{!}{%
\begin{tabular}{lcccccc}
\toprule
\textbf{Source} & $n$ & \textbf{EQ-Bench} & \textbf{Mean NR} & \textbf{Late NR} & \textbf{Retain \%} & \textbf{IR$_{60}$} \\
\midrule
Sonnet 4.6 (vanilla) & 32 & 82.0 [80.3, 83.7] & 0.606 [0.542, 0.668] & 0.137 [0.096, 0.180] & 14.6 [10.8, 18.7] & 0.073 [0.053, 0.094] \\
\textbf{Sonnet 4.6 + Pipeline} & 32 & 85.0 [82.5, 87.1] & 0.749 [0.704, 0.791] & 0.340 [0.273, 0.409] & 34.1 [29.2, 38.9] & 0.135 [0.112, 0.159] \\
\addlinespace
Opus 4.6 (vanilla) & 32 & 82.9 [81.7, 84.0] & 0.611 [0.548, 0.670] & 0.185 [0.140, 0.234] & 15.9 [12.8, 19.1] & 0.090 [0.067, 0.112] \\
\textbf{Opus 4.6 + Pipeline} & 32 & 81.8 [78.9, 84.3] & 0.697 [0.644, 0.747] & 0.296 [0.234, 0.363] & 32.8 [27.5, 38.1] & 0.142 [0.112, 0.175] \\
\addlinespace
GPT-5.2 (vanilla) & 32 & 83.8 [82.3, 85.3] & 0.694 [0.623, 0.756] & 0.329 [0.260, 0.403] & 36.1 [30.7, 41.2] & 0.131 [0.108, 0.153] \\
\textbf{GPT-5.2 + Pipeline} & 32 & 81.2 [78.8, 83.4] & 0.740 [0.677, 0.796] & 0.443 [0.375, 0.512] & 43.6 [38.5, 48.3] & 0.141 [0.113, 0.170] \\
\midrule
Top-10 LLMs (0-shot) & 319 & 81.9 [81.3, 82.4] & 0.630 [0.609, 0.650] & 0.215 [0.197, 0.233] & 23.3 [21.7, 25.0] & 0.106 [0.098, 0.114] \\
\emph{New Yorker} & 261 & 78.6 [75.6, 81.3] & 0.765 [0.746, 0.783] & 0.607 [0.579, 0.633] & 51.7 [49.5, 53.9] & 0.200 [0.189, 0.211] \\
\bottomrule
\end{tabular}%
}
\end{table}

\begin{table}[!t]
\centering
\small
\caption{Paired pipeline-minus-vanilla differences with 95\% bootstrap intervals over the 32 shared prompts. \checkmark\ marks intervals excluding zero.}
\label{tab:ci-diffs}
\resizebox{\textwidth}{!}{%
\begin{tabular}{lcccc}
\toprule
\textbf{Model} & \textbf{Mean NR} & \textbf{Late NR} & \textbf{Retain \%} & \textbf{EQ-Bench} \\
\midrule
Sonnet 4.6 & $+0.143$ [$+0.068$, $+0.221$] \checkmark & $+0.203$ [$+0.122$, $+0.285$] \checkmark & $+19.5$ [$+13.1$, $+25.6$] \checkmark & $+2.99$ [$-0.04$, $+5.68$] n.s. \\
Opus 4.6   & $+0.086$ [$+0.005$, $+0.166$] \checkmark & $+0.111$ [$+0.031$, $+0.191$] \checkmark & $+16.9$ [$+10.7$, $+22.9$] \checkmark & $-1.07$ [$-4.24$, $+1.73$] n.s. \\
GPT-5.2    & $+0.045$ [$-0.042$, $+0.135$] n.s.       & $+0.114$ [$+0.013$, $+0.210$] \checkmark & $+7.5$ [$+0.2$, $+14.7$] \checkmark  & $-2.56$ [$-5.42$, $+0.14$] n.s. \\
\bottomrule
\end{tabular}%
}
\end{table}

\paragraph{What the intervals support.}
Table~\ref{tab:ci-diffs} gives the paired differences. The pipeline's gain on late-stage no-rate, the dimension where LLM stories most clearly lag professional fiction, is significant for all three models, as is the gain on post-spike retention. Mean no-rate improves significantly for Sonnet~4.6 and Opus~4.6 but not for GPT-5.2, whose interval spans zero; we therefore scope the mean-no-rate claim in Section~\ref{sec5} to the two models where it holds. EQ-Bench differences are not significant in either direction for any model, including the $+2.99$ that puts Sonnet~4.6 at the top of the leaderboard. That null is consistent with what the pipeline is designed to do: it targets narrative tension and leaves rubric-style quality where it was.

\paragraph{Comparison against professional fiction.}
Read against the \emph{New Yorker} row, no pipeline configuration reaches professional levels on the discriminating metric. GPT-5.2 with the pipeline attains the highest model mean no-rate at 0.740 [0.677, 0.796], overlapping the \emph{New Yorker} interval of 0.765 [0.746, 0.783], but its late-stage no-rate of 0.443 [0.375, 0.512] remains far below 0.607 [0.579, 0.633], and the two intervals do not overlap. The gap that the metric was built to detect is intact.

\section{Confounders: Length and Genre}
\label{app:confounders}

A no-rate gap that traced back to how long or what kind of story a corpus contains, rather than to how the story is told, would undercut the central claim. This section tests both.

\subsection{Story Length}
\label{app:confounders-length}
\label{app:length-invariance}

Longer stories might mechanically inflate no-rate simply by leaving more plot to diverge on. We test this three ways, measuring length as both sentence count and word count, since the metric is computed per sentence but reported for corpora usually described in words. Both measures give the same answer.

\paragraph{Corpus lengths and within-corpus correlations.}
Table~\ref{tab:length-corr} gives the length distribution of each corpus and the within-corpus correlation between length and mean no-rate. The correlation is positive and significant in \emph{New Yorker} and Tell Me a Story, and indistinguishable from zero in StoryStar and the LLM corpus. Longer stories do tend to score slightly higher where the effect appears at all, which is why we control for it below rather than assert invariance.

\begin{table}[!t]
\centering
\small
\setlength{\tabcolsep}{4pt}
\caption{Length distributions and within-corpus correlations with mean no-rate.}
\label{tab:length-corr}
\resizebox{\textwidth}{!}{%
\begin{tabular}{lccccc}
\toprule
\textbf{Corpus} & $n$ & \textbf{Mean NR} & \textbf{Sentences} & \textbf{Words} & \textbf{$r$(NR, sent) / $r$(NR, words)} \\
\midrule
\emph{New Yorker} & 261 & 0.765 & $241 \pm 114$ & $3365 \pm 1127$ & $+0.290$ ($p{=}1.9{\times}10^{-6}$) / $+0.172$ ($p{=}0.005$) \\
Tell Me a Story & 100 & 0.693 & $115 \pm 45$ & $1411 \pm 518$ & $+0.343$ ($p{=}5{\times}10^{-4}$) / $+0.283$ ($p{=}0.004$) \\
StoryStar & 100 & 0.656 & $217 \pm 123$ & $2612 \pm 1110$ & $-0.009$ ($p{=}0.93$) / $+0.115$ ($p{=}0.25$) \\
Top-10 LLMs & 320 & 0.630 & $144 \pm 67$ & $1367 \pm 602$ & $-0.050$ ($p{=}0.38$) / $-0.065$ ($p{=}0.25$) \\
\bottomrule
\end{tabular}%
}
\end{table}

\paragraph{1. Length does not order the corpora.}
StoryStar is the longer corpus on both measures (217 sentences and 2{,}612 words, against 115 and 1{,}411 for Tell Me a Story) and scores lower on mean no-rate, 0.656 against 0.693. If length drove the ranking, that pair would be reversed.

\paragraph{2. The ranking survives statistical control.}
We fit an OLS regression of mean no-rate on standardized length plus corpus fixed effects, with \emph{New Yorker} as the reference category ($n = 781$ stories with both length measures). Under either length measure, every corpus still sits below \emph{New Yorker} once length is accounted for, and length's own contribution is about $+0.02$ in no-rate per standard deviation (Table~\ref{tab:length-ols}). Tell Me a Story is the one cell that weakens under the word-count control, dropping to $p \approx 0.12$.

\begin{table}[h]
\centering
\small
\caption{OLS regression of mean no-rate on standardized length and corpus ($n = 781$). \emph{Left}: sentences as the length measure ($R^2 = 0.113$; length alone $R^2 = 0.042$; $\Delta R^2$ from corpus $= 0.071$). \emph{Right}: words as the length measure ($R^2 = 0.109$; length alone $R^2 = 0.068$; $\Delta R^2$ from corpus $= 0.041$).}
\label{tab:length-ols}
\begin{tabular}{lcccc}
\toprule
\textbf{Term} & \textbf{Coef} & \textbf{Std err} & \textbf{$t$} & \textbf{$p$} \\
\midrule
\multicolumn{5}{@{}l}{\emph{Sentences as length}}\\
Intercept (\emph{New Yorker}) & $+0.753$ & 0.012 & $+65.5$ & $<10^{-300}$ \\
Sentences (per $+1$ SD)       & $+0.021$ & 0.007 & $+2.99$ & $2.9{\times}10^{-3}$ \\
Tell Me a Story               & $-0.046$ & 0.022 & $-2.08$ & $3.8{\times}10^{-2}$ \\
StoryStar                     & $-0.104$ & 0.021 & $-5.05$ & $5.5{\times}10^{-7}$ \\
Top-10 LLMs                   & $-0.115$ & 0.016 & $-7.22$ & $1.3{\times}10^{-12}$ \\
\midrule
\multicolumn{5}{@{}l}{\emph{Words as length}}\\
Intercept (\emph{New Yorker}) & $+0.745$ & 0.014 & $+54.7$ & $<10^{-200}$ \\
Words (per $+1$ SD)           & $+0.022$ & 0.009 & $+2.40$ & $1.7{\times}10^{-2}$ \\
Tell Me a Story               & $-0.038$ & 0.025 & $-1.54$ & $1.2{\times}10^{-1}$ \\
StoryStar                     & $-0.096$ & 0.021 & $-4.51$ & $7.5{\times}10^{-6}$ \\
Top-10 LLMs                   & $-0.101$ & 0.020 & $-4.94$ & $9.7{\times}10^{-7}$ \\
\bottomrule
\end{tabular}
\end{table}

\paragraph{3. The ranking holds inside every length quartile.}
Tables~\ref{tab:length-quartile-mean} and~\ref{tab:length-quartile-late} bin all stories into quartiles under each length measure and report mean and late-stage no-rate per corpus within the bin. Late-stage no-rate gives the cleanest result: \emph{New Yorker} tops every quartile and the Top-10 LLMs are lowest in every quartile, under both binnings. On mean no-rate, \emph{New Yorker} leads every sentence quartile and word quartiles Q2--Q4; the one exception, word-Q1, contains only 6 \emph{New Yorker} stories, since a \emph{New Yorker} story under 1{,}070 words is atypical. The two remaining metrics behave the same way. Post-spike retention puts Top-10 LLMs last in every quartile of both binnings, with \emph{New Yorker} leading 3 of 4. Inflection rate puts Top-10 LLMs last in Q1--Q3 of both binnings, with \emph{New Yorker} on top in most quartiles.

\begin{table}[h]
\centering
\small
\caption{Mean no-rate within length quartiles (per-corpus mean, $n$ in parentheses). Bins are computed over the pooled set, so a corpus contributes unevenly across quartiles.}
\label{tab:length-quartile-mean}
\begin{tabular}{llcccc}
\toprule
\textbf{Binning} & \textbf{Quartile} & \emph{\textbf{New Yorker}} & \textbf{Tell Me a Story} & \textbf{StoryStar} & \textbf{Top-10 LLMs} \\
\midrule
\multirow{4}{*}{Sentences}
 & Q1 (8--103)     & 0.664 (33)  & 0.637 (46) & 0.577 (16) & 0.610 (102) \\
 & Q2 (103--147)   & 0.747 (31)  & 0.734 (35) & 0.672 (19) & 0.676 (109) \\
 & Q3 (147--240)   & 0.758 (74)  & 0.767 (18) & 0.709 (28) & 0.599 (75) \\
 & Q4 (240--644)   & 0.800 (123) & 0.514 (1)  & 0.643 (37) & 0.609 (34) \\
\midrule
\multirow{4}{*}{Words}
 & Q1 (94--1070)     & 0.587 (6)   & 0.640 (33) & 0.560 (7)  & 0.636 (151) \\
 & Q2 (1070--1789)   & 0.706 (26)  & 0.704 (40) & 0.602 (21) & 0.633 (107) \\
 & Q3 (1789--3125)   & 0.759 (69)  & 0.741 (27) & 0.706 (42) & 0.618 (57) \\
 & Q4 (3125--4977)   & 0.784 (160) & --- (0)    & 0.647 (30) & 0.534 (5) \\
\bottomrule
\end{tabular}
\end{table}

\begin{table}[h]
\centering
\small
\caption{Late-stage no-rate within length quartiles. \emph{New Yorker} is highest and Top-10 LLMs lowest in all eight bins.}
\label{tab:length-quartile-late}
\begin{tabular}{llcccc}
\toprule
\textbf{Binning} & \textbf{Quartile} & \emph{\textbf{New Yorker}} & \textbf{Tell Me a Story} & \textbf{StoryStar} & \textbf{Top-10 LLMs} \\
\midrule
\multirow{4}{*}{Sentences}
 & Q1 (8--103)     & 0.323 (33)  & 0.249 (46) & 0.272 (16) & 0.164 (102) \\
 & Q2 (103--147)   & 0.545 (31)  & 0.329 (35) & 0.290 (19) & 0.229 (109) \\
 & Q3 (147--240)   & 0.589 (74)  & 0.437 (18) & 0.399 (28) & 0.220 (75) \\
 & Q4 (240--644)   & 0.709 (123) & 0.238 (1)  & 0.446 (37) & 0.312 (34) \\
\midrule
\multirow{4}{*}{Words}
 & Q1 (94--1070)     & 0.328 (6)   & 0.247 (33) & 0.255 (7)  & 0.177 (151) \\
 & Q2 (1070--1789)   & 0.325 (26)  & 0.303 (40) & 0.202 (21) & 0.222 (107) \\
 & Q3 (1789--3125)   & 0.551 (69)  & 0.400 (27) & 0.385 (42) & 0.295 (57) \\
 & Q4 (3125--4977)   & 0.687 (160) & --- (0)    & 0.512 (30) & 0.307 (5) \\
\bottomrule
\end{tabular}
\end{table}

\paragraph{Within the \emph{New Yorker} corpus alone.}
Partitioning the 261 \emph{New Yorker} stories by word count gives the same picture from the other direction: predictability is roughly flat across the length range the corpus covers (Table~\ref{tab:length-invariance}), which is what we would expect if no-rate captures the density of narrative surprise per unit of story rather than the amount of story remaining.

\begin{table}[h]
\centering
\small
\caption{Mean and median no-rate by story length for the full \emph{New Yorker} corpus ($n = 261$).}
\label{tab:length-invariance}
\begin{tabular}{lrrr}
\toprule
\textbf{Length bucket} & \textbf{$n$} & \textbf{Mean no-rate} & \textbf{Median no-rate} \\
\midrule
\quad $<$ 2000w   & 41 & 0.696 & 0.723 \\
\quad 2000--3000w & 53 & 0.760 & 0.794 \\
\quad 3000--4000w & 72 & 0.798 & 0.834 \\
\quad 4000--5000w & 95 & 0.772 & 0.784 \\
\bottomrule
\end{tabular}
\end{table}

\paragraph{Scope.}
The length effect inside our range is real but small, and it does not reorder the corpora. We make no claim beyond that range. Every corpus we evaluate sits at or under 5{,}000 words and 640 sentences, and a positive length coefficient of $+0.02$ per standard deviation could compound at novella or novel length in ways this design cannot test. Appendix~\ref{app:cost} takes up what happens to the metric itself at those lengths.

\subsection{Genre}
\label{app:confounders-genre}

On the model side genre is controlled by construction: each of the 32 EQ-Bench prompts fixes a genre, so both the vanilla-versus-pipeline and the Top-10 comparisons are already within-genre. We therefore test genre where it is free to vary, in the \emph{New Yorker} corpus, using the genre labels shipped with the dataset metadata, which follow the macro-genre categories of \citet{williams2017screenwriters}. Restricting to stories at or under 5{,}000 words leaves 242 stories across 10 genres with $n \geq 5$ (Table~\ref{tab:genre}).

\begin{table}[h]
\centering
\small
\caption{Mean no-rate by \emph{New Yorker} macro-genre ($n \geq 5$), against the Top-10 LLM mean of 0.630. One-way ANOVA across genres: $F = 0.61$, $p = 0.79$, $\eta^2 = 2.8\%$.}
\label{tab:genre}
\begin{tabular}{lccc}
\toprule
\textbf{Genre} & $n$ & \textbf{Mean no-rate} & \textbf{vs.\ LLM mean} \\
\midrule
Showbiz/Artistry           & 7  & 0.842 & $+0.212$ \\
School Stories             & 5  & 0.816 & $+0.186$ \\
Psychological              & 17 & 0.786 & $+0.156$ \\
Identity                   & 21 & 0.770 & $+0.140$ \\
Family                     & 55 & 0.765 & $+0.135$ \\
Biography/Character Study  & 56 & 0.759 & $+0.129$ \\
Love                       & 24 & 0.746 & $+0.116$ \\
Addiction                  & 6  & 0.698 & $+0.068$ \\
Ghost/Spirits/Angels       & 5  & 0.680 & $+0.050$ \\
\bottomrule
\end{tabular}
\end{table}

Genre explains 2.8\% of the within-\emph{New Yorker} variance in no-rate and the ANOVA is far from significant. Every well-populated genre sits well above the Top-10 LLM mean, so the corpus-level \emph{New Yorker} $>$ LLM gap of roughly $0.13$ is broad-based across genres rather than driven by a particular high-tension subset, and genre is not a confound for our main results.

\paragraph{Why we do not report a genre-by-genre comparison.}
A direct per-genre comparison between \emph{New Yorker} and LLM stories would be more informative, but the two sets are not labelled under a shared taxonomy. \emph{New Yorker} uses subject-themed macro-genres (Family, Psychological, Love) while EQ-Bench uses writing-style categories (epistolary, folk horror, and so on), and per-genre counts run from 5 to 56 on our side and around 10 on theirs. Any full genre table would be comparing different kinds of label at low power. Where the taxonomies plausibly meet we can match pairs. Pooling EQ-Bench's four romance-flavoured prompt categories against the \emph{New Yorker} Love genre gives 0.746 ($n=24$) against 0.564 ($n=40$), a gap of $+0.182$ (Welch $t = 3.84$, $p = 3.4{\times}10^{-4}$; Mann--Whitney $U = 735$, $p = 4.2{\times}10^{-4}$). The horror pairing is too small to read, at 0.680 ($n=5$) against 0.665 ($n=30$). Across all 22 EQ-Bench categories, LLM no-rate otherwise sits at or below the \emph{New Yorker} mean; the three exceptions (whalepunk fantasy, epistolary, and SCP-style entries) are niche prompts whose structure rather than quality plausibly inflates unpredictability. A properly powered per-genre study needs a shared labelling scheme and more stories per cell, which we leave to future work.

\section{Diversity of the 100 Endings}
\label{app:ending-diversity}

The 100 endings at each position are independent generations (\textsc{Qwen3-32B}, $T{=}1.2$), sampled separately rather than drawn as a single ranked list from one model call, so no ending is conditioned on any other. If they were near-duplicates the metric would collapse, since a story would be scored against one prediction repeated a hundred times. Two checks confirm they are not.

\paragraph{Lexical diversity.}
Table~\ref{tab:lexical-diversity} reports distinct-1, distinct-2, and mean pairwise Jaccard similarity computed across up to 30 stories per group at 8 positions per story. distinct-2 and the low pairwise Jaccard indicate the 100 endings are heterogeneous, not near-duplicates. Crucially, no-rate is only weakly related to lexical diversity across groups (no-rate vs.\ Jaccard $r = -0.21$; vs.\ distinct-2 $r = +0.31$): the six model conditions have nearly identical lexical-diversity profiles yet span the full no-rate range, so no-rate is not a proxy for surface lexical variety but a reflection of plot and event divergence judged against the true ending.

\begin{table}[h]
\centering
\small
\caption{Lexical diversity of the 100 generated endings. Lower pairwise Jaccard means more diverse.}
\label{tab:lexical-diversity}
\begin{tabular}{lccc}
\toprule
\textbf{Group} & \textbf{distinct-1} & \textbf{distinct-2} & \textbf{Mean pairwise Jaccard} \\
\midrule
\emph{New Yorker}      & 0.19 & 0.62 & 0.143 \\
Tell Me a Story        & 0.18 & 0.60 & 0.150 \\
StoryStar              & 0.15 & 0.53 & 0.190 \\
Model conditions (6)   & 0.16--0.18 & 0.59--0.60 & 0.147--0.155 \\
\bottomrule
\end{tabular}
\end{table}

\paragraph{Judged diversity, with anchoring controls.}
Lexical measures can miss semantic near-duplication, so we also had two judges from different model families rate the diversity of the 100-ending sets for 200 \emph{New Yorker} and 200 Top-10 stories on a 1--5 scale (Table~\ref{tab:judged-diversity}). The two judges calibrate differently but agree on the comparison that matters: \emph{New Yorker} and Top-10 ending sets are equally diverse, while their no-rate differs by 0.135. The gap between professional fiction and LLM stories is therefore not about how varied the predictions are, but about how tightly the story constrains where it can go.

\begin{table}[h]
\centering
\small
\caption{LLM-judged diversity of the 100-ending sets (1--5), and anchoring controls. Controls are run on synthetic ending sets: 100 copies of one ending, a 50/50 mix of duplicates and real endings, the real 100 endings, and 100 endings drawn from different stories.}
\label{tab:judged-diversity}
\begin{tabular}{lccc}
\toprule
\textbf{Condition} & $n$ & \textbf{DeepSeek V4 Pro} & \textbf{Gemini 2.5 Flash} \\
\midrule
\emph{New Yorker}                & 200 & $3.04 \pm 0.81$ & $3.92 \pm 0.35$ \\
Top-10 LLMs                      & 200 & $2.98 \pm 0.80$ & $3.92 \pm 0.28$ \\
\midrule
\multicolumn{4}{@{}l}{\emph{Anchoring controls}}\\
100 identical copies             & --- & $1.00 \pm 0.00$ & $1.00 \pm 0.00$ \\
50 duplicates $+$ 50 real        & --- & $1.87 \pm 0.92$ & $1.87 \pm 1.19$ \\
Real 100 endings                 & --- & $3.20 \pm 0.68$ & $3.93 \pm 0.46$ \\
100 endings, different stories   & --- & $4.79 \pm 0.43$ & $4.20 \pm 0.41$ \\
\bottomrule
\end{tabular}
\end{table}

The controls also show that the judges are reading their input rather than emitting a default score: they floor at 1.00 on identical copies, land near the middle on a half-duplicated set, and reserve the top of the scale for endings drawn from different stories. Real ending sets sit between the duplicate floor and the cross-story ceiling, which is what a diverse but story-anchored set should look like.

\section{Computational Cost and Scope}
\label{app:cost}

\paragraph{Measured cost.}
Table~\ref{tab:cost} reports cost measured from the actual runs behind this paper, covering roughly 2{,}050 stories and 97{,}000 judged positions on NVIDIA H200 GPUs. Evaluating one story costs about three GPU-minutes. The low per-position figure comes from prefix caching: the story prefix is shared across all 100 endings at a position, so it is encoded once rather than a hundred times.

\begin{table}[h]
\centering
\small
\caption{Measured cost of the \textit{100-Endings} metric over $\sim$2{,}050 stories and $\sim$97k judged positions.}
\label{tab:cost}
\begin{tabular}{lccc}
\toprule
\textbf{Stage} & \textbf{GPU-hours} & \textbf{Per story (median)} & \textbf{Per sentence position} \\
\midrule
Generation (Qwen3-32B) & $\sim$61 & $\sim$1.8 GPU-min & $\sim$2.3 GPU-s \\
Judge (Qwen3-8B)      & $\sim$37 & $\sim$1.1 GPU-min & $\sim$1.4 GPU-s \\
\midrule
\textbf{Total}                               & $\sim$\textbf{99} & $\sim$\textbf{2.9 GPU-min} & $\sim$\textbf{3.7 GPU-s} \\
\bottomrule
\end{tabular}
\end{table}

\paragraph{Scaling.}
Cost scales linearly in sentences $\times$ 100: a story of $S$ sentences contributes roughly $0.9S$ judged positions after the position filters, each costing a fixed 100 generations and 100 verdicts, so a story ten times longer costs about ten times as much. There is no super-linear term. Within the short-fiction range we report on ($\leq$5{,}000 words), the metric is cheap enough at scale and the cost scales linearly.

\paragraph{Where the metric stops.}
Two ceilings are worth stating plainly. The practical one is context: the generator is served with \texttt{max\_model\_len} of 8{,}192 tokens, so past roughly 5{,}000--6{,}000 English words the story prefix begins to crowd out room for the 100 endings.

The conceptual ceiling is more interesting. Even with unlimited context, asking a model to predict the final ending of a 10{,}000-sentence novel from its first 50 sentences is close to asking for an unconditional base rate. We would expect no-rate to saturate near 1.0 across most early positions and lose its ability to discriminate. The metric is designed for short fiction, where each sentence non-trivially constrains the ending, and we have not validated that the construct survives at novel length. A natural variant would redefine the target at the chapter level, producing one \textit{100-Endings} curve per chapter, which we leave to future work.

\section{Pipeline Ablation}
\label{app:ablation}

The pipeline has three stages, and Steps~1 and~2 are tightly coupled, since Step~2 conditions on the beat plan that Step~1 produces. Removing Step~1 alone would therefore also change Step~2's input rather than isolating a single stage. The informative single-step removal is Step~0, the warmup analysis, which sits upstream of both. We reran Sonnet~4.6 with Step~0 disabled and Steps~1--2 intact, on the same 32 EQ-Bench v3 prompts, and evaluated the outputs with the unchanged \textit{100-Endings} and Qwen3-8B judge stack (Table~\ref{tab:ablation-step0}), then compared conditions prompt by prompt (Table~\ref{tab:ablation-contrasts}). The vanilla and full-pipeline rows are taken directly from Table~\ref{tab:pipeline-100ending}.

\begin{table}[h]
\centering
\small
\caption{Removing Step~0 (warmup analysis) from the Sonnet~4.6 pipeline, on the same 32 EQ-Bench prompts.}
\label{tab:ablation-step0}
\begin{tabular}{lccccc}
\toprule
\textbf{Condition} & $n$ & \textbf{Mean NR} & $\pm$\textbf{SD} & \textbf{Late NR} & $\pm$\textbf{SD} \\
\midrule
Sonnet 4.6, vanilla                  & 32 & 0.606 & 0.184 & 0.137 & 0.125 \\
Sonnet 4.6 $+$ full pipeline         & 32 & \textbf{0.749} & 0.126 & \textbf{0.340} & 0.202 \\
Sonnet 4.6 $+$ pipeline, Step 0 removed & 32 & 0.728 & 0.137 & 0.398 & 0.218 \\
\bottomrule
\end{tabular}
\end{table}

\begin{table}[h]
\centering
\small
\caption{Paired comparisons across the same 32 prompts (paired $t$-test on per-prompt values).}
\label{tab:ablation-contrasts}
\resizebox{\textwidth}{!}{%
\begin{tabular}{lcc}
\toprule
\textbf{Contrast} & \textbf{Mean NR $\Delta$} & \textbf{Late NR $\Delta$} \\
\midrule
Full pipeline $-$ vanilla & $+14.4$pp ($t{=}{+}5.04$, $p{=}4.7{\times}10^{-7}$) \checkmark & $+20.3$pp ($t{=}{+}5.94$, $p{=}2.8{\times}10^{-9}$) \checkmark \\
No-Step-0 $-$ vanilla & $+12.2$pp ($t{=}{+}5.28$, $p{=}1.3{\times}10^{-7}$) \checkmark & $+26.2$pp ($t{=}{+}6.78$, $p{=}1.2{\times}10^{-11}$) \checkmark \\
Full pipeline $-$ No-Step-0 & $+2.2$pp ($t{=}{+}0.79$, $p{=}0.43$) n.s. & $-5.9$pp ($t{=}{-}1.45$, $p{=}0.15$) n.s. \\
\bottomrule
\end{tabular}%
}
\end{table}

\paragraph{Reading.}
Both pipelines beat vanilla decisively on both tension metrics. In terms of the comparison between steps, the bulk of the pipeline's tension gain is attributable to the Steps~1$+$2 (Beat Adaptation $+$ Beat-Conditioned Story Generation) configuration; removing Step~0 produces a slight drop in mean no-rate but an actual gain in late no-rate, and neither difference is statistically detectable. The no-Step-0 EQ-Bench composite of 82.65 also remains inside the full-pipeline interval of [82.5, 87.1] from Table~\ref{tab:ci-table2}, so rubric quality is unchanged as well. In short, the upstream warmup tension-analysis prompt is not load-bearing in this single-iteration setting, and the pipeline's measured advantage over vanilla generation rests on the beat-level plan plus the beat-conditioned generation pass. Practically, the pipeline is more parsimonious than its three-stage description suggests: a re-implementation can drop Step~0 without measurable loss on either axis.

The specific dynamics of combinatorial creativity within this stage (Step~1) will be the focus of the immediate next stage of our future work.

\section{Case Study Details}
\label{app:case_study_stories}

This section provides the detailed analysis and full text of the generated stories discussed in the romance case study (Section~\ref{sec6}, Figure~\ref{fig:bookshop-contrast}). Both stories were generated by Claude Sonnet 4.6 based on the same prompt from the EQ-Bench Creative Writing dataset.

\paragraph{\textit{The New Yorker}.}
In Updike's ``Snowing in Greenwich Village'' (1956), a husband walks an attractive female guest home through the snow while his wife stays behind with a cold. The moment introduces immediate stakes: will the underlying attraction culminate in an affair, or will nothing happen? As Figure~\ref{fig:updike-norate} illustrates, the no-rate spikes once the pair is alone and stays elevated throughout their ambiguous parting at her door. Although the affair never materializes, the narrative sustains the possibility of transgression into the final sentence. Updike maintains the tension to the very end by never allowing the narrative to prematurely settle; the story remains suspended between seduction and restraint, making ambiguity the focal point.

\paragraph{The zero-shot failure.}
Figure~\ref{fig:bookshop-contrast} (left) illustrates the typical failure mode of an unguided LLM attempting a comparable romance, which exhibits a decisive turning point at ~64\% progression. Up to that mark, the banter sustains uncertainty as the owner initially rebuffs the actor. However, the moment she invites him to the back room for tea and acknowledges his honesty, their mutual attraction is explicitly confirmed. With the romantic trajectory resolved, the no-rate drops to $0.29$ and completely flatlines by the final sentences. Unlike Updike, the vanilla model prematurely forecloses the narrative pathway and collapses tension the very moment the characters are alone together.

\paragraph{The pipeline recovery.}
Figure~\ref{fig:bookshop-contrast} (right) shows our pipeline's version of the same prompt. The interaction is driven entirely by subtext, as the characters part without ever acknowledging their underlying chemistry after exchanging only oblique vulnerabilities. This deliberate ambiguity culminates in the story's conclusion, as the distracted owner corrects her ledger: ``She sat with the mistake for a moment before she fixed it.'' Here, \emph{mistake} and \emph{fixed} operate on two levels, referring simultaneously to a mundane accounting error and the unacted-upon romantic spark; consequently, the no-rate remains above 0.85 through 85\% of the story. Like Updike, our pipeline sustains the tension between attraction without resolving into either acceptance or rejection. This highlights that the zero-shot failure pertains to a fundamental inability to \emph{defer closure}, a deficit our pipeline avoids by design through thick planning.

\subsection{EQ-Bench Prompt 03: Love in the Limelight}
\textbf{Genre:} Romance / Slice of Life \\
\textbf{Prompt:} Write a story about a famous actor who ducks into a small Welsh bookshop to hide from the paparazzi. They strike up a conversation with the unimpressed owner.
\begin{itemize}
    \item Include a moment where the actor realizes they are bleeding from a small scratch.
    \item Focus on the contrasting worlds of the two characters.
    \item End with a quiet, ambiguous moment.
\end{itemize}

\subsection{Vanilla Zero-Shot Generation (Claude Sonnet 4.6)}

The bell above the door didn't so much ring as shriek, and Meredith looked up from her inventory sheet to find a man pressed flat against the glass, peering through the gap in the window display like a hunted animal checking for predators.

He was, she noticed with some irritation, absurdly handsome.

``We're open,'' she said. ``If that's what you're wondering.''

He turned. The face resolved itself into something she recognised from cinema screens and magazine covers --- Callum Drake, whose last film had apparently broken three box office records and whose jaw could, according to a review she'd skimmed, ``stop traffic on the M4.'' He looked at her with the particular expression of someone accustomed to being recognised, waiting for the machinery of that recognition to produce its usual effect.

She looked back down at her inventory sheet.

``Brilliant,'' he said, moving away from the door. He was breathing slightly hard. ``I'll just --- browse, shall I?''

``That is generally what one does in a bookshop.''

He laughed --- a short, surprised sound. ``Right. Yes.''

She could feel him hovering. She added three copies of a Patrick O'Brian to the spreadsheet.

``There are about six of them out there,'' he said, from somewhere near Military History. ``Photographers. Someone must have tipped them off I was in Abergavenny.''

``Mm.''

``It's a nice town.''

``I know. I live here.''

Another pause. She heard him actually pull a book from the shelf, which she hadn't expected.

``\textit{The Worst Journey in the World},'' he read aloud. ``Cheerful.''

``Cherry-Garrard. Antarctic expedition. Three men walked into the polar winter in complete darkness to collect penguin eggs for science.'' She set down her pen. ``Two of them nearly died. All three of them suffered what we'd now recognise as severe psychological trauma. Cherry-Garrard spent the rest of his life wondering if he could have saved Scott.''

Callum Drake looked at the book, then at her. ``You're quite intense for a Tuesday.''

``It's Wednesday.''

He smiled then --- the smile, she suspected, the one that had its own fan accounts. It was genuinely devastating, which annoyed her considerably.

``I'm Callum,'' he said.

``I know who you are.''

``And you are?''

``The person who owns this shop.''

He set the book down and leaned against the shelf with the ease of someone who had never once worried about taking up space. ``Come on. I'm hiding from the paparazzi in your establishment. The least you could do is tell me your name.''

``Meredith.'' She picked up her pen again. ``And I'd argue the least I could do is nothing at all, which is what I was doing before you came in.''

``Meredith.'' He said it as though testing the weight of it. ``Do you treat all your customers like this?''

``Only the ones who use my shop as a bolt-hole without asking.''

``Would you have said yes if I'd asked?''

She considered this honestly. ``Probably. But you didn't.''

He had the grace to look slightly abashed, which she hadn't anticipated. The charm, she was realising, had a kind of autopilot quality --- it ran continuously, like background software --- but underneath it something else flickered when it failed to land.

``Fair point,'' he said quietly.

She softened, fractionally, against her better judgement. ``There's tea in the back. If you want to wait properly.''

He followed her through the curtain into the small room behind the counter --- two chairs, an electric kettle, towers of unshelved stock. She filled the kettle without looking at him.

``Do you ever get tired of it?'' she asked. ``People recognising you everywhere.''

``Tired of it?'' He sat down in one of the chairs. ``God, no. I'd be terrified if they stopped.'' He said it lightly, but the lightness had a careful quality, like a lid on something.

She turned to look at him properly for the first time. ``That's an honest answer.''

``Don't tell anyone. It'll ruin the mystique.'' He looked at his hands. ``My dad was an actor. Fringe theatre, mostly --- never made it. I used to watch him at the dinner table, just \textit{performing}, even at home. Telling stories, making everyone laugh. And I thought that was just --- that was just who he was.'' He stopped. ``Took me until I was about thirty to understand he was terrified of the room going quiet. That if nobody was watching, he didn't quite know what he was.''

The kettle began to build toward boiling.

``And you?'' Meredith said.

``Me?''

``Are you terrified of the room going quiet?''

He looked at her for a long moment. Outside, distantly, she heard a car pass on the high street.

``I haven't let it go quiet long enough to find out,'' he said.

She poured the water. She was very aware, suddenly, of the smallness of the room, the particular quality of the afternoon light coming through the single high window, the fact that her hands were not entirely steady.

``Milk?'' she asked.

``Please.'' Then: ``You're not going to tell me I'm brilliant and it must be so wonderful, are you.''

``No.''

``Most people do.''

``I'm sure they do.''

``It's---'' He stopped again. Started differently. ``It's quite refreshing, actually. Talking to someone who's not---''

``Performing back at you?''

He looked startled. Then he laughed --- not the polished one, but something rougher and more real. ``Yeah,'' he said. ``Exactly that.''

She handed him his tea and sat down in the opposite chair with her own. Through the curtain, the shop waited --- its shelves of difficult journeys and long winters and people who had nearly not survived the thing they'd chosen.

``The photographers will probably give up in an hour,'' she said.

``Probably,'' he agreed.

Neither of them suggested he leave sooner than that.

\subsection{Narrative Pipeline Generation (Claude Sonnet 4.6)}

The bell didn't ring; it clattered, a harsh, broken sound that made Elara look up from her ledger. 

The man who shoved his way inside was soaked, his trench coat plastered to his shoulders. He threw his weight against the door, pressing his forehead against the frosted glass. Outside, the flash of cameras strobed through the rain, turning the quiet Welsh street into a silent, frantic disco.

``Sorry,'' the man said, breathless. 

``We're closed.''

``Just five minutes.'' He didn't turn around. He was still watching the street, his shoulders tense. 

Elara didn't move from the counter. She watched a drop of water fall from the hem of his coat and soak into the threadbare rug. ``You're dripping on the folios.''

He finally turned. The ambient light from the streetlamp caught his face. It was a face built for fifty-foot screens, all sharp angles and calculated symmetry, currently ruined by exhaustion and the frantic energy of a hunted animal. He reached up to push his wet hair back and stopped. 

His fingers came away red. He stared at the blood, then touched a scratch on his cheek, right below his eye. 

``It's raining,'' he said, as if that explained everything.

``I hadn't noticed.''

He let out a short, hollow laugh and wiped the blood on his coat sleeve. He started to pace the narrow aisle, running a hand over the spines of the books. His movements were too large for the space, too dramatic, as if he were trying to hit a mark on a stage.

``Do you have any poetry? Contemporary?''

``Second aisle, back wall.''

He stopped pacing and pulled a slim volume from the shelf. He didn't open it. He just held it, staring at the cover. 

``I don't read poetry,'' he said to the spine of a Doris Lessing novel. ``But someone told me I should start.''

``Usually people start with the classics.''

``I don't have time for the classics.''

``I don't think they mind.''

He looked at her then, really looked at her. The manic energy drained out of him, leaving something brittle in its place. 

``Do you know what it's like,'' he said, ``to spend six months pretending to be someone else, and then when you stop, you aren't sure if you exist when no one is watching?''

The silence stretched. Outside, the flashes stopped. The paparazzi were moving on. 

Elara closed her ledger. She picked up a pen and began to trace the grain of the wood on the counter. 

``I had a cat,'' she said. ``A ginger tom. He died three years ago. Sometimes I still step over the place where he used to sleep on the rug.''

He didn't move. He didn't say anything. He just watched her trace the wood grain. 

``Did you get another cat?''

``No.''

He nodded, slowly. He walked to the counter and set the book down. ``I'll take the poetry.''

``That's \pounds4.50.''

He pulled a damp ten-pound note from his pocket and laid it on the wood. ``Keep the change.''

``We don't do tips.'' She slid the change across the counter. 

He didn't take it immediately. His hand hovered over the coins, close to hers. The air in the shop felt suddenly thin. Then he scooped the money up, put the book in his pocket, and walked out into the rain. 

Elara watched him go. She picked up her pen and opened the ledger. She wrote the sale in the wrong column, putting the \pounds4.50 under expenses instead of income. 

She sat with the mistake for a moment before she fixed it.

\end{document}